\documentclass{article} 
\usepackage{iclr2027_conference,times}

\usepackage{amsmath,amsfonts,bm}

\def\eqref#1{equation~\ref{#1}}

\def\1{\bm{1}}

\DeclareMathAlphabet{\mathsfit}{\encodingdefault}{\sfdefault}{m}{sl}
\SetMathAlphabet{\mathsfit}{bold}{\encodingdefault}{\sfdefault}{bx}{n}

\usepackage{hyperref}
\usepackage{url}
\usepackage{url}            
\usepackage{booktabs}       
\usepackage{amsfonts}       
\usepackage{nicefrac}       
\usepackage{microtype}      
\usepackage{xcolor}         
\usepackage{makecell}
\usepackage{multirow}
\usepackage{graphicx}
\usepackage{comment}
\usepackage{bbding}
\usepackage[table]{xcolor}
\usepackage{amsmath} 
\usepackage{algorithm,algorithmic}
\usepackage{graphicx}
\usepackage{booktabs}
\usepackage{hyperref}
\usepackage{listings}
\usepackage{xspace}

\title{Restoring without Forgetting: Filter-Level Continual Image Restoration via Parameter-Space Integrated Gradients}

\author{Xin Feng$^1$, Jin Zhao$^2$, Yizhen Zhang$^3$, Wenjie Pei$^4$, Fanglin Chen$^4$, Guangming Lu$^4$ \\
$^1$School of Informatics, University of Edinburgh \quad $^2$Baidu Inc. \\
$^3$Tsinghua University \quad $^4$Harbin Institute of Technology, Shenzhen
}

\iclrfinalcopy 
\begin{document}

\maketitle

\begin{abstract}
Adapting image restoration models to a stream of new tasks without revisiting past data remains challenging due to catastrophic forgetting. In this work, we propose Restoring without Forgetting (RwF), a filter-level continual adaptation framework for image restoration built upon a critical observation: task-specific knowledge is centered in a small subset of filters and can be separated from those reconstructing general content. RwF first performs parameter-space integrated gradients attribution to localize degradation-critical filters in a coarse-to-fine manner. It then adapts to new tasks by generating task-specific filters from a filter bank using compact factorized low-rank transformations, further augmented with cross-task attention and prototypical contrastive learning, and lastly assembles them back only at localized positions. Experiments on six restoration tasks show that RwF effectively avoids forgetting and achieves competitive restoration quality against all-in-one methods that have full data access, and outperforms LoRA-style adaptation with $\sim$10$\times$ fewer additional parameters. Code is available at {\hypersetup{hidelinks}\href{https://github.com/funkdub/Restoring-without-Forgetting}{https://github.com/funkdub/RwF}}.
\end{abstract}

\vspace{-8pt}
\section{Introduction}
\label{sec:intro}
\vspace{-8pt}

Recent image restoration methods are evolving from the single-task specialist model~\citep{mansour2023zero, chen2023masked, kong2023efficient,wang2023multi, gao2024efficient, dong2015image} to unified backbones~\citep{jiang2024autodir,cui2025adair,potlapalli2023promptir} that handle multiple degradations simultaneously.
Nevertheless, real-world deployments rarely encounter a fixed and closed set of degradations: new corruption types, capture conditions, and device characteristics emerge over time. In many practical scenarios, revisiting historical training data is infeasible due to storage, privacy, or data ownership constraints. As a result, naively fine-tuning a restoration model on each incoming task often leads to catastrophic forgetting~\citep{goodfellow2013empirical}, naturally framing the problem as continual image restoration. 

While continual learning has been extensively studied for high-level recognition~\citep{kirkpatrick2017overcoming,rebuffi2017icarl,qu2021recent,joseph2022energy,sun2023regularizing}, extending it to image restoration is more challenging. First, restoration is evaluated at the pixel level, where even mild forgetting can cause visible artifacts and sharp drops in terms of metrics. Second, restoration models usually learn degradation patterns implicitly, and these patterns are entangled with image content, thus generic regularization often struggles to preserve task-specific performance.
Existing replay-based methods~\citep{rebuffi2017icarl,lopez2017gradient,wang2022continual,shin2017continual,cong2020gan} require storing samples from previous tasks, which is often impractical when access to past data is limited, and it also increases training cost over time. Architectural expansion approaches~\citep{rusu2016progressive,serra2018overcoming,hung2019compacting,kang2022forget} can protect past knowledge but often incur significant parameter growth when applied to modern restoration networks. Parameter-efficient adaptation methods, like Adapter~\citep{houlsby2019parameter,chen2022vision} or LoRA~\citep{hu2022lora}, reduce per-task overhead, yet they generally inject trainable parameters broadly and do not explicitly identify which parts of the backbone are truly responsible for modeling each degradation, limiting both scalability and robustness under sequential adaptation.

We address these limitations with Restoring without Forgetting (RwF), a filter-level continual adaptation framework inspired by a critical observation~\citep{xie2021finding,sundararajan2016gradients}: degradation-specific knowledge is centered in a small subset of filters and can be separated from filters that primarily reconstruct general content (detailed in Appendix~\ref{app:observe}). Concretely, RwF first performs parameter-space integrated gradients attribution, and localizes degradation-critical filters through integrated gradient values in a coarse-to-fine manner. To efficiently acquire new restoration capabilities, RwF maintains a filter bank of previously localized filters and generates task-specific filters using compact factorized low-rank transformations, which are then assembled back only at the localized positions. We further introduce cross-task attention to reweight historical knowledge during filter generation, and perform prototypical contrastive learning to promote task-separated representations.
Our contributions are summarized as follows:
\vspace{-4pt}
\begin{itemize}
    \setlength{\itemsep}{0pt}
    \item We propose a filter-level continual adaptation framework to achieve Restoring without Forgetting (RwF) for continual image restoration under a sequential setting where past-task data cannot be revisited, effectively avoiding catastrophic forgetting.
    \item We introduce integrated gradients in parameter space to identify and retain a small set of degradation-critical filters, then generate task-specific filters from a filter bank via factorized low-rank transforms.
    \item Extensive experiments on six restoration tasks demonstrate that RwF significantly avoids forgetting, achieves competitive quality compared with all-in-one restoration models, and outperforms continual learning methods and LoRA-style adaptation while requiring about an order of magnitude fewer additional parameters.
\end{itemize}

\begin{figure*}[!t]
    \centering
    \vspace{-16pt}
    \includegraphics[width=1\textwidth]{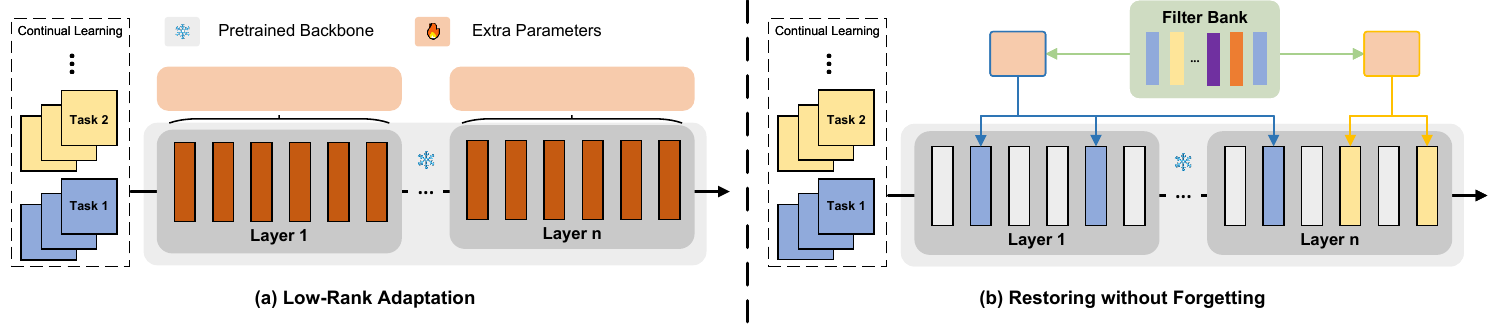}
    \vspace{-16pt}
    \caption{(a) LoRA updates every filter of each adapted layer through low-rank matrices. (b) RwF localizes the few degradation-critical filters of each task and regenerates only them from a shared filter bank, leaving the rest of the backbone frozen.}
    \label{fig:pipeline}
    \vspace{-8pt}
\end{figure*}

\vspace{-12pt}
\section{Related Work}
\label{sec:rw}

\noindent\textbf{Image Restoration} has progressed from task-specific models~\citep{mansour2023zero,chen2023masked,kong2023efficient,wang2023multi,gao2024efficient,dong2015image} to generic backbones~\citep{wang2022uformer,feng2023u2,zamir2021multi} and all-in-one models~\citep{potlapalli2023promptir,jiang2024autodir,cui2025adair} that handle multiple degradations at once, yet all assume joint training with every degradation available upfront. When corruptions arrive sequentially in the continual learning setting, fine-tuning on each catastrophically forgets the others.

\noindent\textbf{Continual Learning} learns a sequence of tasks without catastrophic forgetting.
\textit{Regularization-based} methods restrict updates to parameters important for past tasks~\citep{kirkpatrick2017overcoming,zeng2019continual} or distill previous outputs~\citep{li2017learning} but weaken over long task streams~\citep{hsu2018re,farquhar2018towards}; \textit{replay-based} methods rehearse stored or generated samples~\citep{rebuffi2017icarl,lopez2017gradient,wang2022continual,shin2017continual,cong2020gan} at the cost of computation and privacy; \textit{architecture-based} methods isolate task-specific parameters~\citep{rusu2016progressive,serra2018overcoming,hung2019compacting,kang2022forget} at the cost of scalability.
These methods are mainly designed for classification tasks, where a discrete label absorbs feature drift, but restoration demands pixel-level fidelity, so drift surfaces as visible artifacts, motivating adaptation confined to the parameters each task relies on.

\noindent\textbf{Parameter-Efficient Fine-Tuning} methods such as adapters~\citep{houlsby2019parameter,chen2022vision} and LoRA~\citep{hu2022lora} bound per-task overhead with few added parameters, but place them uniformly across layers, blind to which components encode a degradation. Continual generation methods~\citep{zhai2020piggyback,zhai2021hyper} factorize filters into shared and task-specific parts yet fix the split per layer, and LIRA~\citep{liu2020lira} adds a degradation-specific branch per task with distillation and GAN-based pseudo-replay, paying for both expansion and replay. In contrast, we localize the filters each degradation relies on via Integrated Gradients~\citep{sundararajan2017axiomatic,sundararajan2016gradients} and adapt only those, separating content reconstruction from degradation-specific behavior without extra branches or replay.

\section{Method}
\subsection{Overview}
In this work, we propose an integrated gradient-based filter learning approach to solve the problem of continual learning for image restoration models. The main objective of this problem is to train a model $\mathcal{M}$ parameterized by $\theta$ that can handle a series of image restoration tasks $\{\mathcal{T}_t\}_{t=1}^{T}$, constrained by the condition that the learning process can only access one task at a time and training data for each task is available only once.

Formally, an incoming training sample of task $t$ can be represented as $\mathcal{D}_t=\{\mathcal{X}_t,\mathcal{Y}_t\}$, herein $\mathcal{X}_t$ is the input degraded images, $\mathcal{Y}_t$ is the corresponding groundtruth, $t\in\mathcal{T}=\{1,...,T\}$ is the index of image restoration tasks. The objective function of this problem can be summarized as 
\vspace{-4pt}
\begin{equation}
    \min_{\theta} \sum^T_{t=1} \mathbb{E}_{(\mathcal{X}_t,\mathcal{Y}_t)}\left\|\mathcal{M}(\mathcal{X}_t)-\mathcal{Y}_t\right\|_1.
\end{equation}
\vspace{-8pt}

\begin{figure*}[!t]
    \centering
    \vspace{-32pt}
    \includegraphics[width=1\textwidth]{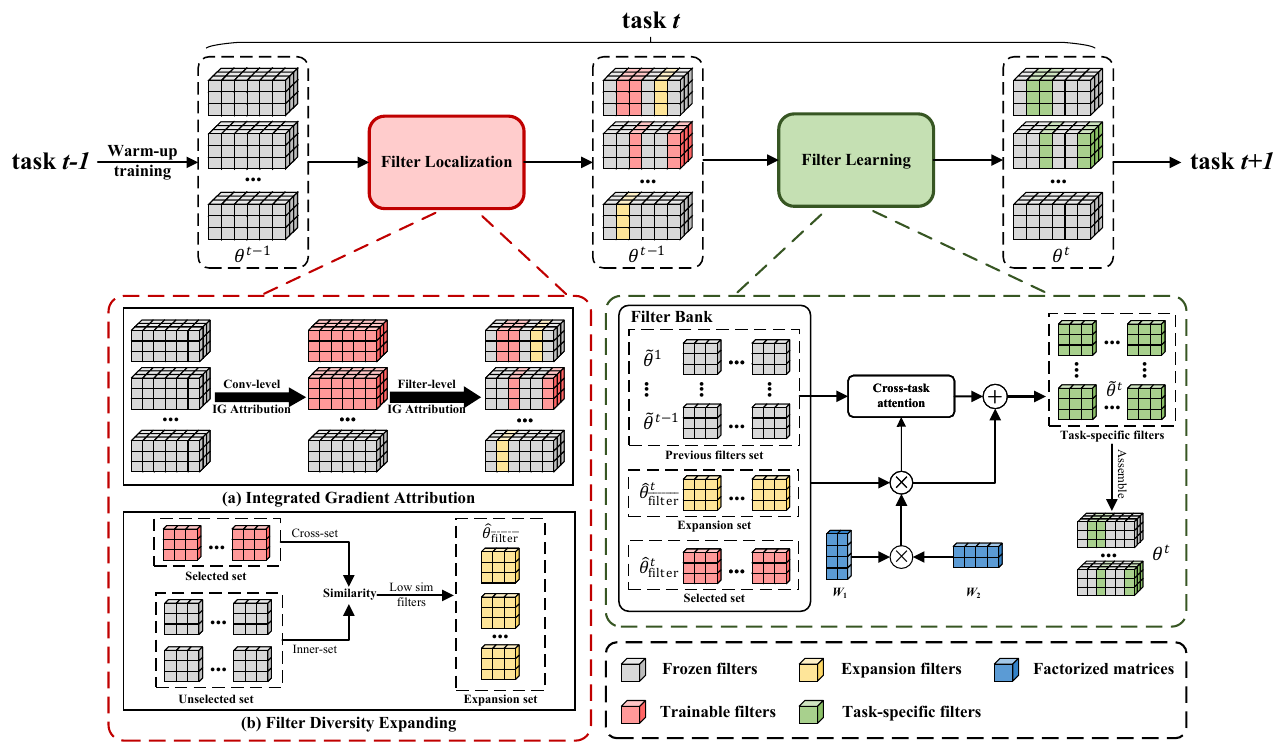}
    \vspace{-20pt}
    \caption{Overview of RwF. Filter localization (left): parameter-space integrated gradients rank layers, then filters within them, and the selected set is expanded with high-IG, low-similarity filters. Filter learning (right): the localized filters are regenerated from a filter bank by factorized matrices with cross-task attention and assembled back into the frozen backbone.}
    \label{fig:pipeline}
    \vspace{-12pt}
\end{figure*}

As presented in Fig.~\ref{fig:pipeline}, our RwF model mainly consists of two phases, the filter localization phase and filter learning phase. In the filter localization phase, the restoration model $\mathcal{M}$ is warmed up on the new task for 1k iterations and leverages the integrated gradient method to find the key filters that contribute to modeling the degradation patterns. The localized key filters are employed to serve as prior knowledge to learn new degradation patterns for next task of image restoration. In the filter learning phase, the restoration model $\mathcal{M}$ is able to efficiently perform continual learning by maintaining a filter bank and only updating a small number of key filters for new tasks. 

\subsection{Adaptive filter localization for task-specific degradation}
\smallskip\noindent\textbf{Revisiting the Integrated Gradient (IG).}
IG~\citep{sundararajan2017axiomatic,sundararajan2016gradients} is proposed to quantify pixel-wise importance and identify features essential for accurate image recognition. 
Given a classifier \(F:\mathbb{R}^n \to [0,1]\), an input image \(x \in \mathbb{R}^n\), and a black baseline image \(\bar{x} \in \mathbb{R}^n\), the (vanilla) integrated gradients are defined as the path integral of the gradients along the straight-line path from \(\bar{x}\) to \(x\):
\begin{equation}
\mathrm{IG}_i\!\left(x\right)
= (x_i - \bar{x}_i)
\int_{\alpha=0}^{1}
\frac{\partial F\!\big(\bar{x} + \alpha\times(x - \bar{x})\big)}{\partial x_i}\,\mathrm{d}\alpha ,
\label{equ:1}
\end{equation}
where $x_i$ denotes the value of $i$-th pixel and $\frac{\partial F}{\partial x}$ is the gradient of $F$. Thus, the accumulated gradients are able to reflect the importance of each pixel.

Inspired by previous research in image super-resolution~\citep{xie2021finding}, the integrated gradient has great potential to find key filters across diverse restoration tasks, thus addressing specific degradations by cumulating filter gradients. 
Imitating the way of integrated gradient for pixels, we first train a baseline network $\overline{F}$ that only learns to reconstruct the input image itself rather than restoring any degradation. In the meanwhile, a target network $F$ is trained to restore a specific task of image restoration.
Building upon the contrast between these two networks, we design a two-step filter localization strategy via integrated gradient.

\smallskip\noindent\textbf{Integrated gradient-based task-specific filters localization.}
In order to find the key filters that are essential to restore degradation patterns of a specific restoration task, we accumulate the integrated gradients for all filters by the intermediate model between the baseline model $\overline{F}$ parameterized by $\overline{\theta}$ and target restoration model $F$. The intermediate model is obtained by $\gamma(\alpha)=\alpha\overline{\theta} + (1-\alpha)\theta$. Thus, Equ.~\ref{equ:1} can be approximated by discrete states of intermediate models:
\begin{equation}
    \text{IG}_i(\theta) \approx \frac{1}{N}(\theta-\overline{\theta})_i\sum^N_{k=1}\left[\frac{\partial\mathcal{L}(\gamma(\alpha),x)}{\partial\gamma(\alpha)}\bigg|_{\gamma(\alpha),\alpha=k/N}\right],
    \label{equ:2}
\end{equation}
where $N=50$ is the number of discrete states.
Consequently, integrated gradients effectively measure the importance of each filter for a specific restoration task.

Following Equ.~\ref{equ:2}, we aim to localize the key filters $\theta_{\text{filter}}$ for the current restoration task $t$ in a coarse-to-fine manner. Specifically, we coarsely obtain conv-level integrated gradients of all filters to localize the key convolutional layers by
\begin{equation}
    \hat{\theta}_{\text{conv}} = \mathbb{I}_{\text{IG}(\theta_{\text{conv}})>P_k}(\theta),
    \label{equ:3}
\end{equation}
where $P_k$ is the percentile, $\text{IG}(\theta_{\text{conv}})=\sum \text{IG}(\theta_{\text{filter}})$, $\hat{\theta}_{\text{conv}}=\{\hat{\theta}_{\text{filter1}},\hat{\theta}_{\text{filter2}},...\}$ denotes the selected convolutions with higher IG values. 
This coarse step allows to coarsely locate the convolutional layers that primarily contribute to degradation restoration rather than basic content reconstruction. Subsequently, within these selected layers, we re-evaluate the integrated gradients at the filter level to precisely identify the key filters and pinpoint their exact locations.
\begin{equation}
    \hat{\theta}_{\text{filter}}, L= \mathbb{I}_{\text{IG}(\hat{\theta}_{\text{filter}})>P_k}(\hat{\theta}_{\text{conv}}), 
    \label{equ:4}
\end{equation}
where $P_k$ is the percentile, $\hat{\theta}_{\text{filter}}$ denotes the parameters of preliminarily selected filters with higher IG values, $L=\{l_1,l_2,\dots,l_r\}$ is the corresponding locations of $r$ selected key filters. Note that the selected filters are only a small fraction of the full restoration model. 
Therefore, to prevent catastrophic forgetting, our approach explicitly retains the identified filters $\hat{\theta}_{\text{filter}}$. Building upon this precise localization for task $t$, our continual filter learning mechanism repurposes this minimal filter subset to efficiently adapt to new restoration tasks.

\smallskip\noindent\textbf{Adaptive IG threshold.}
In filter localization stage described in Equ.~\ref{equ:3} and Equ.\ref{equ:4}, it adopts a percentile $P_k$ to select key filters. However, different restoration tasks exhibit varying levels of complexity, and the distribution of integrated gradient scores can differ substantially across tasks. A fixed threshold may lead to either excessive or insufficient selection of key filters. Thus, we introduce an adaptive thresholding mechanism that dynamically adjusts the percentile based on the statistical properties of the IG score distribution. Specifically, we compute the variance $\sigma^2$ and the information entropy $\mathcal{E}$ of IG scores across all filters:
\vspace{-4pt}
\begin{equation}
    \mathcal{E} = -\sum_{i} \hat{g}_i \log(\hat{g}_i + \epsilon),
    \label{equ:entropy}\vspace{-4pt}
\end{equation}
where $\hat{g}_i$ is the normalized IG score of the $i$-th filter and $\epsilon$ is a small constant. Based on $\sigma^2$ and $\mathcal{E}$, the adaptive percentile is determined by:
\begin{equation}
    P_k^{\text{adapt}} = \text{clip}\left(P_k^{\text{base}} + \lambda_\sigma \cdot \sigma^2 - \lambda_\mathcal{E} \cdot \mathcal{E},\; P_{\min},\; P_{\max}\right),
    \label{equ:adaptive_pk}
\end{equation}
where $P_k^{\text{base}}$ is the base percentile, $\lambda_\sigma$=0.05 and $\lambda_\mathcal{E}$=0.05 are scaling factors, and $[P_{\min}, P_{\max}]$ defines the valid range (e.g., [85, 98]). 
Intuitively, when the IG scores are highly concentrated (low variance and low entropy), a higher percentile is used to select only the most critical filters; when the scores are dispersed, a lower percentile is adopted to retain more candidate filters for subsequent diversity expanding. This enables our model to cope with degradation tasks of varying complexity in a self-adaptive manner.

\smallskip\noindent\textbf{Filter diversity expanding.}
One potential problem is that the selected key filters $\hat{\theta}_{\text{filter}}$ in Equ.~\ref{equ:4} might be similar. 
To relieve it, we design a filter diversity expanding strategy which supplements unselected filters $\hat{\theta}_{\overline{\text{filter}}}$ by Equ.~\ref{equ:4} with higher integrated gradients as well as lower similarity with $\hat{\theta}_{\text{filter}}$. 
Thus, it can provide adequate prior knowledge to strengthen the performance of subsequent filter learning.
Specifically, the filter diversity expanding strategy first calculates the cosine similarity between $\hat{\theta}_{\text{filter}}$ and $\hat{\theta}_{\overline{\text{filter}}}$ by
\begin{equation}
\small
\begin{split}
    \mathcal{S}_{j} &= \frac{1}{|\hat{\theta}_{\text{filter}}|}\sum_{i}\cos(\hat{\theta}_i,\hat{\theta}_j)+\frac{1}{|\hat{\theta}_{\overline{\text{filter}}}|}\sum_{k}\cos(\hat{\theta}_j,\hat{\theta}_k)\\
    &=\frac{1}{|\hat{\theta}_{\text{filter}}|}\sum_{i}\frac{\hat{\theta}_i\cdot\hat{\theta}_j}{\Arrowvert\hat{\theta}_i\Arrowvert~\Arrowvert\hat{\theta}_j\Arrowvert}+\frac{1}{|\hat{\theta}_{\overline{\text{filter}}}|}\sum_{k}\frac{\hat{\theta}_j\cdot\hat{\theta}_k}{\Arrowvert\hat{\theta}_j\Arrowvert~\Arrowvert\hat{\theta}_k\Arrowvert},
    \label{equ:5}
\end{split}   
\end{equation}
where $\hat{\theta}_i\in \hat{\theta}_{\text{filter}}$, $\hat{\theta}_j$ and $\hat{\theta}_k\in \hat{\theta}_{\overline{\text{filter}}}$, and $\mathcal{S}$ denotes the similarity score. 
Then, it selects filters with the top-$k$ lowest similarity scores from $\hat{\theta}_{\overline{\text{filter}}}$ and supplements them to generate the final set of key filters $\theta_{\text{filter}}$ for new task.

\subsection{Continual Filter Learning Mechanism}
After filter localization, we propose a continual filter learning mechanism to seamlessly adapt restoration model to new tasks (see Fig.~\ref{fig:pipeline}), which enables restoration model to continuously learn new task $t$ by optimizing the localized key filters, $\hat{\theta}_{\text{filter}}$. 

\smallskip\noindent\textbf{Continual filter learning.}
Rather than training independent models for individual restoration tasks, our continual filter learning mechanism optimizes only the transformation matrices $W$ to update the filters $\theta_{\text{filter}}$ for each new task. Furthermore, we maintain a filter bank to archive previously learned key filters, which serve as prior knowledge for the subsequent continual learning process. Prior to the formal training phase, the model must first trace back to its previous state at $\overline{F}$. Notably, this approach incurs tiny parameter overhead while effectively avoiding catastrophic forgetting by preserving task-specific key filters.

Specifically, for a new task $t$, the filter learning process (illustrated in Fig.~\ref{fig:pipeline}) is divided into two steps: filter generation and model assembly. Formally, the filter generation process is defined as:
\vspace{-2pt}
\begin{equation}
    \tilde{\theta}^t=\left[\mathcal{R}^{-1}(\mathcal{R}([\tilde{\theta}_{\text{filter}}^1, \tilde{\theta}_{\text{filter}}^2,\dots,\hat{\theta}_{\text{filter}}^{t-1}, \hat{\theta}_{\overline{\text{filter}}}^t])\cdot W_1^t\cdot W_2^t), \hat{\theta}_{\text{filter}}^t\right],
    \label{equ:6}
\end{equation}
where $\mathcal{R}$ and $\mathcal{R}^{-1}$ denote the reshape operation and its inverse, respectively. $\tilde{\theta}^t$ represents the task-specific filters generated for task $t$. To enhance parameter efficiency and reduce redundancy, transformation matrix $W$ is factorized into two matrices, $W_1^t$ and $W_2^t$.

Following the generation of task-specific filters, the model assembly is formulated as:
\vspace{-4pt}
\begin{equation}
        \theta^t=\mathcal{G}(\theta^{t-1},\tilde{\theta}^t,L^t),
\end{equation}
where $\theta^t$ denotes complete model parameters for task $t$, $\mathcal{G}$ symbolizes the assembly operator, and $L^t$ indicates the index of key filters localized for task $t$. More precisely, the assembly is computed via:
\begin{equation}
    \theta_i^t=\sum_{j\notin L^t}\delta_{ij}\theta_i^{t-1}+\sum_{1\le k\le r^t}\delta_{il_k^t}\tilde{\theta}_k^t,
\end{equation}
\begin{equation}
    \delta_{ij}=
    \begin{cases}
        1 & \text{if } i=j, \\
        0 & \text{if } i\neq j,
    \end{cases}
    \label{equ:7}
\end{equation}
where $\delta_{ij}$ is the Kronecker delta, and $\theta_i^t$ represents the $i$-th filter of the updated model. By decoupling task-specific adaptation from the backbone, our continual filter learning mechanism enables the model to accommodate novel restoration tasks with high computational efficiency. Consequently, the proposed framework effectively eliminates catastrophic forgetting while maintaining a minimal parameter overhead throughout the continual learning process.

\smallskip\noindent\textbf{Cross-task attention for filter generation.}
While Equ.~\ref{equ:6} generates task-specific filters by matrix multiplication over filter bank, it treats all historical filters equally without considering the varying relevance of different prior tasks to the current task. To selectively leverage and transfer the most relevant prior knowledge during filter generation, we introduce a lightweight cross-task attention mechanism.
Specifically, we treat the currently generated parameter $\tilde{\theta}^t_{\text{cur}}$ as the query, and the historical task parameters $\{\hat{\theta}_{\text{filter}}^i\}_{i=1}^{t-1}$ preserved in the filter bank as keys and values. The cross-task attention computes a weighted aggregation of historical knowledge:
\begin{equation}
    \text{Att}(Q, K, V) = \text{Softmax}\left(\frac{Q K^\top}{\sqrt{d}}\right) V,
    \label{equ:cross_attn}
\end{equation}
where $Q = W_Q \tilde{\theta}^t_{\text{cur}}$, $K = W_K [\hat{\theta}_{\text{filter}}^1, \dots, \hat{\theta}_{\text{filter}}^{t-1}]$, $V = W_V [\hat{\theta}_{\text{filter}}^1, \dots, \hat{\theta}_{\text{filter}}^{t-1}]$, and $W_Q, W_K, W_V$ are lightweight linear projections with hidden dimension $d$. The attended historical knowledge is then injected into the current filter through a residual connection:
\begin{equation}
    \tilde{\theta}^t = \tilde{\theta}^t_{\text{cur}} + \gamma \cdot \text{Att}(Q, K, V),
    \label{equ:cross_residual}
\end{equation}
where $\gamma$=0.1 is a scaling factor to ensure the stability of filter generation process. 

\subsection{Loss Functions}
\label{sec:loss}

To optimize our proposed RwF model of previously learned restoration capabilities, we employ a composite loss function comprising three components: pixel reconstruction loss, prototypical contrastive loss, and frequency-domain loss.

\smallskip\noindent\textbf{Pixel reconstruction loss.}
Following the typical way of pixel-level supervision for image restoration, we employ the $\ell_1$ pixel reconstruction loss to minimize the distance between the restored image $\mathcal{M}(\mathcal{X}_t)$ and the corresponding groundtruth $\mathcal{Y}_t$ for task $t$:
    $\mathcal{L}_{\text{pixel}} = \left\|\mathcal{M}(\mathcal{X}_t) - \mathcal{Y}_t\right\|_1.$

\smallskip\noindent\textbf{Prototypical contrastive loss.}
A key challenge in continual learning is to maintain discriminative feature representations across sequentially learned tasks. To this end, we design a prototypical contrastive loss that encourages the feature representations to form distinguishable clusters for different restoration tasks in the latent feature space. Specifically, we maintain a set of prototypes of feature embeddings $\{p_i\}_{i=1}^{t}$, each of which serves as the representative embedding for a corresponding degradation task $i$.

For current task $t$, we extract latent embeddings $e$ from the intermediate layers of restoration model $\mathcal{M}$. Prototypical contrastive loss pulls $e$ towards the current task prototype $p_t$ while pushing it away from all historical task prototypes $\{p_i\}_{i=1}^{t-1}$:
\begin{equation}
\mathcal{L}_{\text{proto}} =
-
\log
\frac{
\exp\left(-\|e - p_t\|_2^2 / \tau \right)
}{
\sum_{i=1}^{K}
\exp\left(-\|e - p_i\|_2^2 / \tau \right)
},
\label{eq:proto}
\end{equation}
where $K$ is the number of existing tasks and $\tau$=0.07 is the temperature coefficient.
In this way, restoration model can maintain a discriminative latent feature space and precisely model diverse degradation patterns across sequential training. 
To maintain the stability of prototypes during continuous training, we update each prototype using an Exponential Moving Average (EMA):
\begin{equation}
    p_t \leftarrow \mu \cdot p_t + (1 - \mu) \cdot \overline{e}_t,
    \label{equ:momentum}
\end{equation}
where $\mu$=0.9 is a momentum coefficient and $\overline{e}_t = \frac{1}{|\mathcal{B}|}\sum_{f \in \mathcal{B}} f$ denotes the mean embeddings computed over the current mini-batch $\mathcal{B}$. Such momentum update mechanism ensures the prototype evolves smoothly, thereby avoiding the instability caused by noisy gradients in individual batches.

\smallskip\noindent\textbf{Frequency-domain loss.}
Considering that most image restoration tasks demand high-fidelity recovery of high-frequency details, we further introduce a frequency-domain loss based on the Fast Fourier Transform (FFT). Given the restored image $\hat{\mathcal{Y}}_t = \mathcal{M}(\mathcal{X}_t)$ and its groundtruth $\mathcal{Y}_t$, we compute the FFT loss as:
\begin{equation}
    \mathcal{L}_{\text{FFT}} = \left\|\mathcal{F}(\hat{\mathcal{Y}}_t) - \mathcal{F}(\mathcal{Y}_t)\right\|_1 + \lambda_h \left\|\mathcal{H} \odot \left(\mathcal{F}(\hat{\mathcal{Y}}_t) - \mathcal{F}(\mathcal{Y}_t)\right)\right\|_1,
    \label{equ:fft}
\end{equation}
where $\mathcal{F}(\cdot)$ denotes the 2D FFT operation, $\mathcal{H}$ is a high-pass filter that emphasizes high-frequency components, and $\lambda_h$=1.5 is a weighting factor. The second term enforces additional supervision on the high-frequency spectrum. 

The overall loss function for training our RwF model is:
\begin{equation}
    \mathcal{L} = \mathcal{L}_{\text{pixel}} + \alpha \mathcal{L}_{\text{proto}} + \beta \mathcal{L}_{\text{FFT}},
    \label{equ:total}
\end{equation}
where $\alpha$=0.1 and $\beta$=0.01 are hyper-parameters to balance the contributions of different loss terms. 

\newcommand{\red}[1]{\textcolor[rgb]{1,0,0}{#1}}       

\section{Experiments}
\label{sec:experiments}
In this section, we first conduct preliminary experiments on self-synthesized datasets with varied degradations to investigate the effectiveness of each RwF property. 
Then, we evaluate RwF on public restoration benchmarks comprising four diverse restoration tasks with substantially more challenging degradation patterns, and compare against recent parameter-efficient adaptation and all-in-one methods. To demonstrate our method's generality, we equip RwF with {three} distinct backbones: a \textbf{Transformer}-based model {HINT}~\citep{zhou2025devil}, 
a \textbf{Convolutional} model {MPRNet}~\citep{zamir2021multi}, and a \textbf{Diffusion} model {DiffUIR}~\citep{zheng2024selective}.

\smallskip\noindent\textbf{Implementation Details.}
{All experiments employ the AdamW~\citep{loshchilov2017decoupled} optimizer at a constant learning rate, $5\times10^{-4}$ for dehazing and $2\times10^{-4}$ for the remaining stages, with weight decay between $0$ and $1\times10^{-4}$ and $(\beta_1,\beta_2)=(0.9,0.999)$. 
Each task is trained for $150$k iterations with a $1$k-iteration warmup. 
Filter localization uses $50$ integration steps over $8$ mini-batches with an adaptive percentile threshold between $85$ and $88$ at tensor level, followed by diversity expansion. The prototypical term maintains one Exponential-Moving-Average (EMA) latent vector per task rather than a sample buffer.}
We evaluate the quality of restored images in terms of PSNR and SSIM. For more implementation details, please see Appendix~\ref{app:imple}.

\vspace{-4pt}
\subsection{Preliminary Experiments}
\label{sec:preliminary}
\vspace{-4pt}

\begin{figure*}[!t] 
    \centering
    
    \begin{minipage}{0.6\linewidth}
        \centering
        \includegraphics[width=\linewidth]{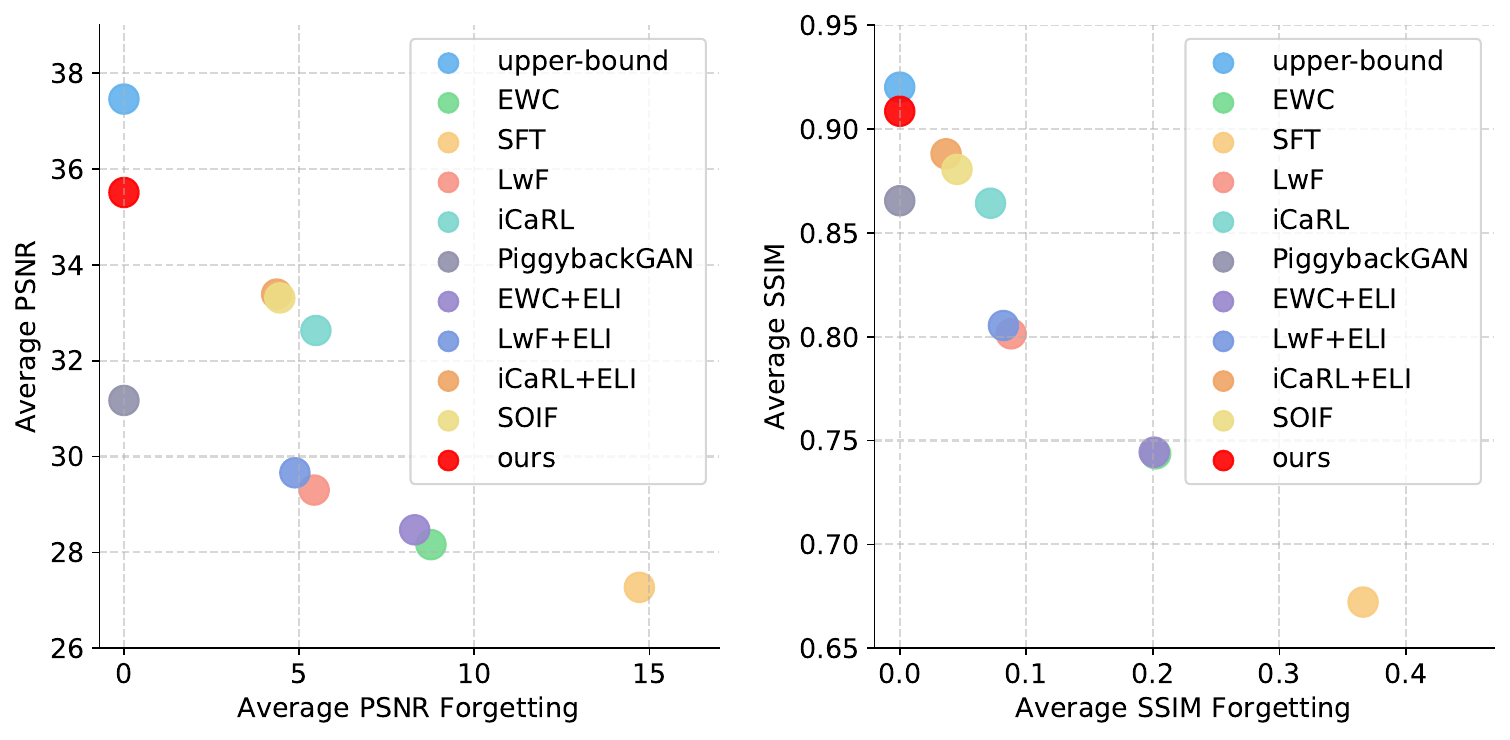}
        \vspace{-8pt}
        \caption{Comparison of forgetting levels for different methods. X-axis stands for the average forgetting, and Y-axis stands for the average metric.}
        \label{fig:bubble}
    \end{minipage}
    \hspace{8pt}
    \begin{minipage}{0.35\linewidth}
        \centering
        \includegraphics[trim=0 0 558pt 0, clip, width=0.82\linewidth]{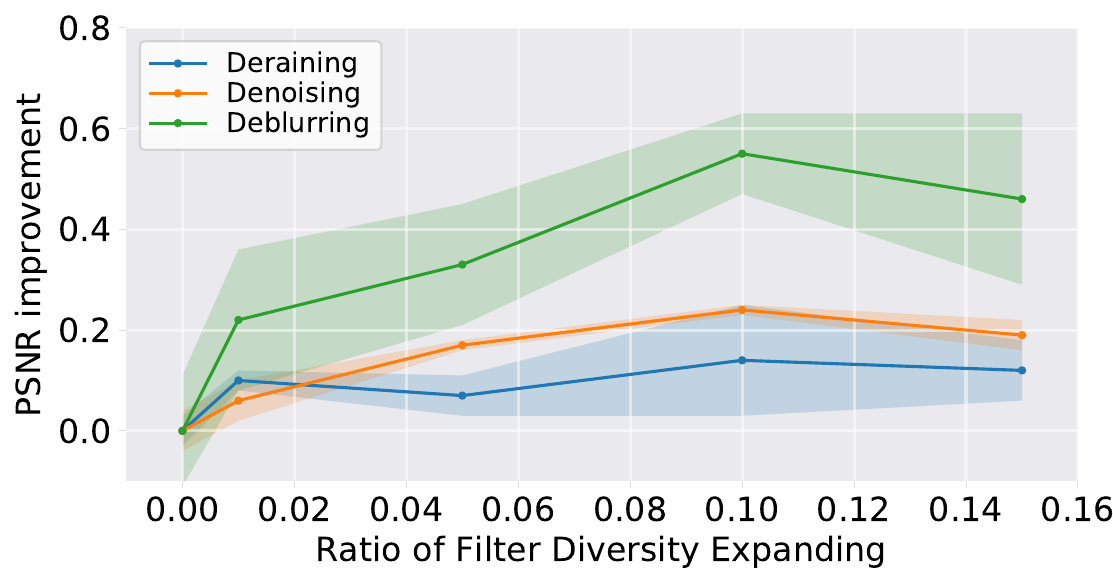}
        
        \vspace{-4pt} 
        
        \includegraphics[trim=536pt 0 0 0, clip, width=0.85\linewidth]{images/fde_ratio.pdf}
        
        \vspace{-8pt}
        \caption{Performance gains with different ratios of filter diversity expanding.}
        \label{fig:dissimilar_ratio}
    \end{minipage}
    
    \vspace{-12pt}
\end{figure*}

We first conduct preliminary experiments on a synthetic dataset from DIV2K~\citep{Agustsson_2017_CVPR_Workshops} with diverse degradation patterns, including rain, noise, and blurriness, in order to investigate the effectiveness of our RwF for continual image restoration. Methods are evaluated on all tasks after learning the final task. More experimental details are in Appendix~\ref{app:imple}.

\begin{figure}[!t]
    \centering
    \includegraphics[width=0.8\columnwidth]{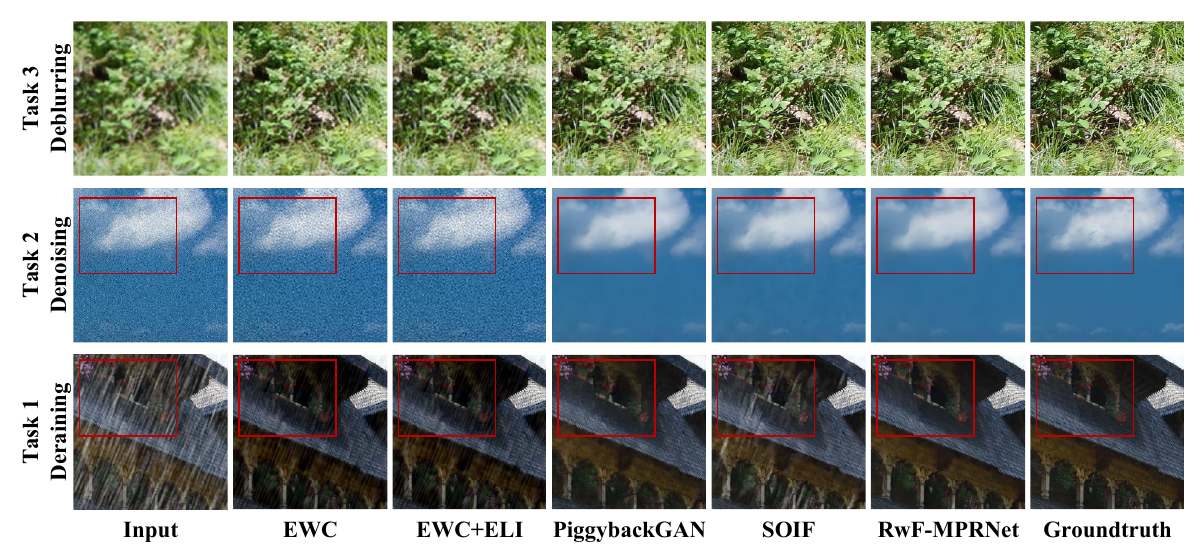}
    \vspace{-4pt}
    \caption{Visualization of restored images on three randomly selected samples from test set. {Bounding boxes emphasize remaining degradation artifacts resulting from catastrophic forgetting.}}
    \label{fig:visualization_comparing}
    \vspace{-8pt}
\end{figure}

As shown in Tab.~\ref{table:table1}, our RwF-equipped MPRNet~\citep{zamir2021multi} achieves the best restoration performance on the two previously learned tasks against typical continual learning baselines. Compared to PiggybackGAN, our RwF achieves comparable restoration quality (Fig.~\ref{fig:bubble}). Qualitative comparisons in Fig.~\ref{fig:visualization_comparing} further show better preservation of image structures and textures across tasks.

\vspace{-8pt}
\subsection{Ablation Study}
\vspace{-4pt}

\smallskip\noindent\textbf{Effect of adaptive task-specific filters localization.} 
Our task-specific filter localization process is decoupled into three progressive stages: {Integrated Gradient Attribution (IGA), Convolution-level Localization (CL), and Filter-level Localization (FL).} Note that the baseline is LoRA finetuning with the same parameter amount, disabling all our proposed components. As shown in Tab.~\ref{tab:ablation_final}, integrating either CL or FL yields substantial performance gains in continual image restoration over the variant lacking IG attribution. This aligns with the intuition that integrated gradients effectively capture the relative importance of trainable parameters, providing an essential foundation for accommodating new tasks. Moreover, FL consistently outperforms its conv-level counterpart CL, further demonstrating the effectiveness of RwF's coarse-to-fine localization strategy.

\smallskip\noindent\textbf{Effect of continual filter learning mechanism.} 
We conduct experiments to measure the contribution of Filter Bank (FB), Cross-Task Attention (CTA), and $\mathcal{L}_{\text{proto}}$. 
Disabling FB and CTA limits selective knowledge transfer, degrading the model's adaptation to new tasks due to interference from irrelevant historical filters.
Furthermore, removing $\mathcal{L}_{proto}$ degrades average performance since filter-level retention alone is insufficient without maintaining task-discriminative latent representations.
As further shown in Fig.~\ref{fig:dissimilar_ratio}, setting the filter diversity expanding ratio to 10\% provides an optimal trade-off between performance gain and parameter overhead.

\subsection{Experiments on Public Restoration Benchmarks}
\label{sec:pub}

\begin{table*}[!t] 
    \centering
    
    \begin{minipage}{0.67\linewidth}
        \centering
        \caption{Quantitative results after continual learning on three tasks, trained in the order of deraining, denoising, and deblurring.}
        \label{table:table1}
        \vspace{4pt} 
        \resizebox{\linewidth}{!}{
        \begin{tabular}{l|rr rr rr rr}
        \toprule
        \multicolumn{1}{c|}{\multirow{2}{*}{{Method}}} & \multicolumn{2}{c}{{Deraining}} & \multicolumn{2}{c}{{Denoising}} & \multicolumn{2}{c}{{Deblurring}} & \multicolumn{2}{c}{{Average}} \\
        \cmidrule(lr){2-3} \cmidrule(lr){4-5} \cmidrule(lr){6-7} \cmidrule(lr){8-9}
        \multicolumn{1}{c|}{} & PSNR & SSIM & PSNR & SSIM & PSNR & SSIM & PSNR & SSIM \\
        \midrule
        SFT-MPRNet~\citep{zamir2021multi} & 23.16 & 0.648 & 21.48 & 0.437 & \textbf{37.16} & \textbf{0.922} & 27.27 & 0.670 \\
        EWC~\citep{kirkpatrick2017overcoming} & 27.33 & 0.754 & 26.88 & 0.642 & 30.27 & 0.835 & 28.16 & 0.744 \\
        LwF~\citep{li2017learning} & 29.60 & 0.853 & 29.24 & 0.755 & 29.05 & 0.796 & 29.30 & 0.801 \\
        iCaRL~\citep{rebuffi2017icarl} & 28.88 & 0.807 & 32.60 & 0.865 & 36.40 & 0.922 & 32.63 & 0.864 \\
        PiggybackGAN~\citep{zhai2020piggyback} & 32.51 & 0.895 & 29.54 & 0.824 & 31.46 & 0.878 & 31.17 & 0.866 \\
        EWC+ELI~\citep{joseph2022energy} & 28.07 & 0.756 & 27.05 & 0.643 & 30.30 & 0.835 & 28.47 & 0.744 \\
        LwF+ELI & 30.12 & 0.864 & 29.80 & 0.756 & 29.07 & 0.796 & 29.66 & 0.805 \\
        iCaRL+ELI & 30.26 & 0.845 & 33.48 & 0.897 & 36.44 & 0.922 & 33.39 & 0.888 \\
        SOIF~\citep{sun2023regularizing} & 30.40 & 0.846 & 33.10 & 0.875 & 36.42 & 0.921 & 33.31 & 0.881 \\
        RwF-MPRNet (ours) & \textbf{38.52} & \textbf{0.942} & \textbf{33.69} & \textbf{0.874} & {34.49} &{0.913} & \textbf{35.57} & \textbf{0.910} \\
        \bottomrule
        \end{tabular}
        }
    \end{minipage}
    \hspace{4pt}
    \begin{minipage}{0.22\linewidth}
        \centering
        \caption{Ablation study on RwF components. }
        \label{tab:ablation_final}
        \vspace{4pt} 
        \resizebox{\linewidth}{!}{
        \begin{tabular}{l rr}
        \toprule
        \multicolumn{1}{c}{\multirow{2}{*}{\textbf{Variant}}} & \multicolumn{2}{c}{\textbf{Average}} \\
        \cmidrule(l){2-3}
        & PSNR & SSIM \\
        \midrule
        Baseline & 34.64 & 0.896 \\
        \midrule
         w/o CL & 35.32 & 0.906 \\
         w/o FL & 35.18 & 0.904 \\
         w/o FB & 35.12 & 0.901 \\
         w/o CTA & 35.26 & 0.905 \\
         w/o $ \mathcal{L}_{\mathrm{pro}} $ & 35.40 & 0.907 \\
        \midrule
        \textbf{RwF (full)} & \textbf{35.57} & \textbf{0.910} \\
        \bottomrule
        \end{tabular}
        }
    \end{minipage}
    
    \vspace{-12pt}
\end{table*}

To further investigate the potential of RwF for continual image restoration, we conduct experiments on a series of restoration tasks with more challenging and diverse degradation patterns, using both public synthetic and real-world restoration datasets.

\smallskip\noindent\textbf{Experimental Setup.}
We assemble a four-task benchmark from widely used public datasets:
Rain100L~\citep{yang2017deep} (deraining), SOTS~\citep{li2018benchmarking} (dehazing), Snow100K~\citep{liu2018desnownet} (desnowing), and LOLv2-Real~\citep{yang2021sparse} (low-light enhancement).
Their degradation patterns vary across heterogeneous image conditions and exhibit domain-specific distributions. Besides, we further combine our RwF with a Transformer-based restoration backbone HINT~\citep{zhou2025devil} to assess architecture generality. 

\smallskip\noindent\textbf{Baselines.}
{Besides the Sequential Fine-Tuning (SFT) reference,} we introduce two additional categories of methods: (1) \emph{all-in-one restoration methods}: including AdaIR~\citep{cui2025adair}, and AutoDIR~\citep{jiang2024autodir}, which train a single model jointly on all tasks with simultaneous data access; (2) \emph{parameter-efficient fine-tuning}: represented by Low-Rank Adaptation (LoRA)~\citep{hu2022lora}, which freezes the backbone and learns task-specific low-rank adapters for each new task. 

\begin{figure}[!t]
\centering
\includegraphics[width=0.875\columnwidth]{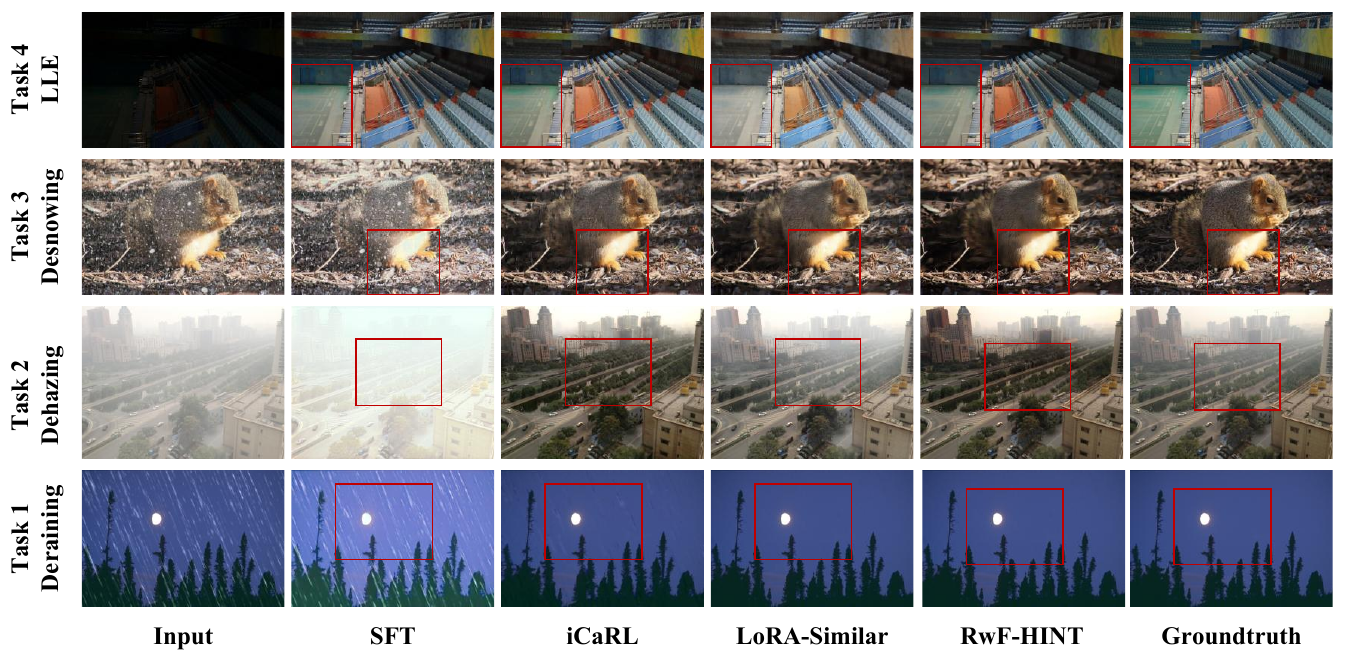}
\vspace{-12pt}
\caption{Visual comparison on the four-task public benchmark after learning the last task (LLE).}
\label{fig:visualization_comparing2}
\vspace{-12pt}
\end{figure}

\begin{table}[!t]
\centering
\caption{Quantitative results on the four-task public restoration benchmark with the order of deraining, dehazing, desnowing, and low-light enhancement (LLE). Best results are in \textbf{bold} and the second are \underline{underlined}. Note that all-in-one methods are not ranked.
}
\resizebox{0.825\linewidth}{!}{
\begin{tabular}{l|ccccccccccc}
\toprule
\multicolumn{1}{c|}{\multirow{2}{*}{\textbf{Method}}} 
& \multicolumn{2}{c}{{Deraining}} 
& \multicolumn{2}{c}{{Dehazing}} 
& \multicolumn{2}{c}{{Desnowing}} 
& \multicolumn{2}{c}{{LLE}}
& \multicolumn{2}{c}{{Average}} & Param \\
\cmidrule(lr){2-3} \cmidrule(lr){4-5} \cmidrule(lr){6-7} \cmidrule(lr){8-9} \cmidrule(lr){10-11}
\multicolumn{1}{c|}{} 
& PSNR & SSIM 
& PSNR & SSIM 
& PSNR & SSIM 
& PSNR & SSIM 
& PSNR & SSIM & increase \\
\midrule

AutoDIR~\citep{jiang2024autodir} 
& 34.49 & 0.961 
& 29.31 & 0.970 
& 27.56 & 0.868 
& 20.54 & 0.799 
& 27.98 & 0.900 & - \\

AdaIR~\citep{cui2025adair} 
& 36.88 & 0.980 
& 30.34 & 0.975 
& 28.95 & 0.901 
& 21.80 & 0.821 
& 29.49 & 0.919 & - \\

\midrule
{SFT-MPRNet}~\citep{zamir2021multi} 
& 17.29 & 0.767 
& 13.89 & 0.712 
& 15.24 & 0.698 
& 22.40 & \underline{0.864} 
& 17.21 & 0.760 & -\\

{SFT-DiffUIR}~\citep{zheng2024selective}
& 23.45 & 0.819 
& 27.39 & 0.940 
& 21.20 & 0.746 
& 21.37 & 0.860 
& 23.35 & 0.841 & -
\\

SFT-HINT~\citep{zhou2025devil} 
& 14.84 & 0.736 
& 10.33 & 0.674 
& 11.75 & 0.574 
& 20.04 & 0.855
& 14.24 & 0.711 & -\\

LwF~\citep{li2017learning} 
& 27.52 & 0.885 
& 19.84 & 0.841 
& 20.33 & 0.782 
& 22.15 & 0.851 
& 22.46 & 0.840 & - \\

iCaRL~\citep{rebuffi2017icarl} 
& 33.11 & 0.950 
& 26.34 & 0.946 
& 25.89 & 0.840 
& \underline{22.68} & 0.856 
& 27.01 & 0.898 & - \\

LoRA-Similar~\citep{hu2022lora} 
& \textbf{38.72} & \textbf{0.984} 
& 27.35 & 0.958 
& 26.34 & 0.858 
& 18.13 & 0.804 
& 27.64 & 0.901 & 0.26M\\

LoRA-Full~\citep{hu2022lora} 
& \textbf{38.72} & \textbf{0.984} 
& 29.69 & \underline{0.973} 
& \underline{27.97} & \textbf{0.878} 
& 20.16 & 0.848 
& 29.14 & \underline{0.921} & 1.73M\\

\textbf{RwF-MPRNet} 
& \underline{36.98} & \underline{0.979} 
& 29.50 & 0.967 
& 27.48 & 0.865 
& 20.04 & 0.838 
& 28.50 & 0.912 & 0.59M \\

\textbf{RwF-DiffUIR} 
& 34.65 & 0.966 
& \textbf{33.56} & \textbf{0.982} 
& \textbf{28.15} & 0.872 
& \textbf{23.50} & \textbf{0.867} 
& \textbf{29.97} & \textbf{0.922} & 0.53M \\

\textbf{RwF-HINT} 
& \textbf{38.72} & \textbf{0.984} 
& \underline{29.73} & \underline{0.973} 
& 27.94 & \underline{0.876} 
& 20.73 & 0.849 
& \underline{29.28} & \underline{0.921} & \textbf{0.17M} \\

\bottomrule
\end{tabular}}
\vspace{-8pt}
\label{table:table8}
\end{table}
\smallskip\noindent\textbf{Comparison with all-in-one methods.}
All-in-one methods access all task data simultaneously, thus their performance should inherently represent the upper bound of continual learning methods. However, we observe that our RwF-HINT model outperforms the all-in-one baseline AutoDIR~\citep{jiang2024autodir} across four restoration tasks by 1.30 dB average PSNR and approaches the performance of the state-of-the-art all-in-one restoration baseline AdaIR~\citep{cui2025adair}. It proves that our RwF framework retains the advantages of sequential training and knowledge assembling.
 
\smallskip\noindent\textbf{Comparison with LoRA.}
While LoRA~\citep{hu2022lora} effectively limits forgetting by constraining adaptation to low-rank subspaces, it has to balance content reconstruction and degradation restoration to train a new task, which obviously limits its performance of continual image restoration: LoRA-Similar trails RwF by 1.64 dB average PSNR on RwF-HINT and LoRA-Full with ten times params than RwF-HINT by 0.14 dB.
By contrast, RwF explicitly models task-specific filters for each new restoration with the guidance of previous knowledge, resulting in competitive performance of continual restoration with limited parameter increase.


\smallskip\noindent\textbf{Qualitative Analysis.}
Fig.~\ref{fig:visualization_comparing2} presents visualized results by competing methods across all four restoration tasks. Typical continual learning baselines, SFT and iCaRL~\citep{rebuffi2017icarl}, produce poor results on learned results due to severe catastrophic forgetting. 
While both LoRA-Similar and our proposed RwF-HINT effectively mitigate forgetting and preserve restoration performance on previously learned tasks, LoRA~\citep{hu2022lora} still exhibits visible artifacts and a noticeable loss of high-frequency details. This degradation stems from its insufficient utilization of learned knowledge from prior tasks. In contrast, RwF consistently maintains high-fidelity, task-specific restoration quality across all tasks, avoiding visible artifacts and over-smoothing issues.

\vspace{-4pt}
\section{Conclusion}
\vspace{-4pt}

In this paper, we present Restoring without Forgetting (RwF), a highly efficient continual learning framework tailored for image restoration. By localizing task-critical filters via adaptive integrated gradients, RwF effectively decouples degradation-specific patterns from general content reconstruction. During sequential adaptation, our method dynamically generates and assembles specialized filters using factorized low-rank transformations guided by a historical filter bank. Extensive evaluations over six restoration tasks across three backbones confirm that RwF optimally mitigates catastrophic forgetting, matching the performance of all-in-one restoration models with only a fraction of the parameter overhead.


\subsection*{AI use statement}
In this work, we used generative AI tools only to check grammar and correct typographical errors in the manuscript text. We did not use generative AI tools to propose the research idea, design the method, implement or run experiments, analyze results, produce figures, or write any of the scientific content. We reviewed every AI-suggested correction before accepting it and take full responsibility for the final content of this work.

\bibliography{iclr2027_conference}
\bibliographystyle{iclr2027_conference}

\newpage
\appendix

\section{Preliminary}
\label{app:preliminary}
Inspired by Integrated Gradients (IG)~\citep{sundararajan2017axiomatic,sundararajan2016gradients}, our RwF adopts a path-integral view of attribution to localize task-specific filters in image restoration models. Since our continual restoration framework ultimately needs to decide which model components should be preserved and which should be adapted, input-space saliency alone is insufficient. Therefore, we first revisit the principle of IG in its original input-space form, and then explain how the same idea can be migrated to the filter level in parameter space.

\subsection{Integrated Gradients in Input Space}
IG~\citep{sundararajan2017axiomatic,sundararajan2016gradients} was originally introduced as an axiomatic attribution framework for deep networks. Consider a network $F: \mathbb{R}^d \to \mathbb{R}$, an input $x \in \mathbb{R}^d$, and a baseline input $\bar{x} \in \mathbb{R}^d$. The baseline represents the absence of informative evidence, e.g., a black image. Instead of attributing the prediction of $F(x)$ using only the local gradient at $x$, IG accumulates gradients along the straight-line path from $\bar{x}$ to $x$:

$$ \mathrm{IG}_{i}(x) = (x_{i}-\bar{x}_{i}) \int_{0}^{1} \frac{\partial F\!\left(\bar{x}+\alpha\times(x-\bar{x})\right)}{\partial x_{i}} \, d\alpha . $$
A key property of IG is the completeness relation, which follows from the chain rule and the fundamental theorem of calculus. Denoting the interpolation path by $\gamma(\alpha) = \bar{x} + \alpha(x - \bar{x})$, we have
$$ \sum_{i=1}^{d}\mathrm{IG}_{i}(x) = \int_{0}^{1} \nabla F(\gamma(\alpha))^{\top} \frac{\partial \gamma(\alpha)}{\partial \alpha} \, d\alpha = F(x)-F(\bar{x}). $$
IG decomposes the total output difference between the target input and the baseline input into additive contributions from individual input dimensions. In other words, attribution is grounded in a path integral of the model’s differential response. In this way, IG attribution is able to identify which input pixels are responsible for the transition from an uninformative baseline input to the final prediction.

Nevertheless, while input-space IG attribution can identify important pixels, it fails to locate task-specific parameters essential for continual image restoration. Since image restoration relies on filters to model complex degradation patterns (e.g., rain or haze), spatial saliency alone cannot pinpoint which model filters should be updated. Furthermore, to ensure selective plasticity and avoid catastrophic forgetting, we shift the attribution target from pixels to filters, thus enabling precise localization and adaptation of task-critical filters during continual learning.

\subsection{From Input-space to Parameter-space Attribution}
In this section, we provide a derivation of the Equ.3 in Sec.3.2 of the main paper.
Let $\theta$ be the parameter of the restoration model where $\theta \in \Theta$, $\Theta \subseteq \mathbb{R}^{m}$ is the whole parameter space, $x \in X$ be the input images where $X \subseteq \mathbb{R}^{n}$. Thus, the loss function $\mathcal{L} : \Theta \times X \rightarrow \mathbb{R}$ can be denoted as $\mathcal{L}(\theta, x)$.
For an input image $x$, we can represent the changes of the model functions by the changes of the loss function:
\begin{equation}
\Delta\mathcal{L} = \mathcal{L}(\theta,x) - \mathcal{L}(\overline{\theta},x) = \sum_{i}\int_{C}\frac{\partial\mathcal{L}(\theta,x)}{\partial\theta_{i}}d\theta_{i}, 
\label{equ:e1}
\end{equation}
where the $\overline{\theta}$ and $\theta$ are the parameters of the baseline model and the target model, respectively. 
$C \subseteq \Theta$ is the path between $\overline{\theta}$ and $\theta$ in parameter space, $\theta \in C$ are all points on $C$ and $\theta_{i}$ is the $i$-th dimension parameter of $\theta$. 
Let $\gamma(\alpha)$, $\alpha \in [0,1]$ be the parameter form of $C$ which satisfies $\gamma(0)=\theta$, $\gamma(1)=\overline{\theta}$. Then, Equ.~\ref{equ:e1} can be rewritten as: 
\begin{equation}
\Delta\mathcal{L} = \mathcal{L}(\gamma(1),x) - \mathcal{L}(\gamma(0),x) = \sum_{i}\int_{0}^{1}\frac{\partial\mathcal{L}(\gamma(\alpha),x)}{\partial\gamma(\alpha)_{i}}\times\frac{\partial\gamma(\alpha)_{i}}{\partial\alpha}d\alpha. 
\end{equation}
Thus, the integrated gradients of the $i$-th dimension parameter could be defined as: 
\begin{equation}
\text{IG}_{i}(\theta) = \int_{0}^{1}\frac{\partial\mathcal{L}(\gamma(\alpha),x)}{\partial\gamma(\alpha)_{i}}\times\frac{\partial\gamma(\alpha)_{i}}{\partial\alpha}d\alpha.
\label{equ:e3}
\end{equation}
A simple and effective choice for the path is the straight line between $\overline{\theta}$ and $\theta$. Then, the path could be represented as $\gamma(\alpha) = \alpha\overline{\theta} + (1-\alpha)\theta$, where $\gamma(0)=\theta$, $\gamma(1)=\overline{\theta}$. Thus, the Equ.~\ref{equ:e3} can be rewritten as: 

\begin{equation}
\text{IG}_{i}(\theta) = (\theta-\overline{\theta})_{i}\int_{0}^{1}\frac{\partial\mathcal{L}(\gamma(\alpha),x)}{\partial\gamma(\alpha)_{i}}d\alpha. 
\end{equation}
Then, the integral can be approximated by discrete states sampled along the path: 
\begin{equation}
\text{IG}_{i}(\theta) \approx \frac{1}{N}(\theta-\overline{\theta})_{i}\sum_{k=1}^{N}\left. \frac{\partial\mathcal{L}(\gamma(\alpha),x)}{\partial\gamma(\alpha)_{i}} \right|_{\alpha=k/N} 
\end{equation}

\section{Localized Filters Carry the Restoration Capability}
\label{app:observe}

\begin{figure*}[!b]
    \centering
    \includegraphics[width=0.85\textwidth]{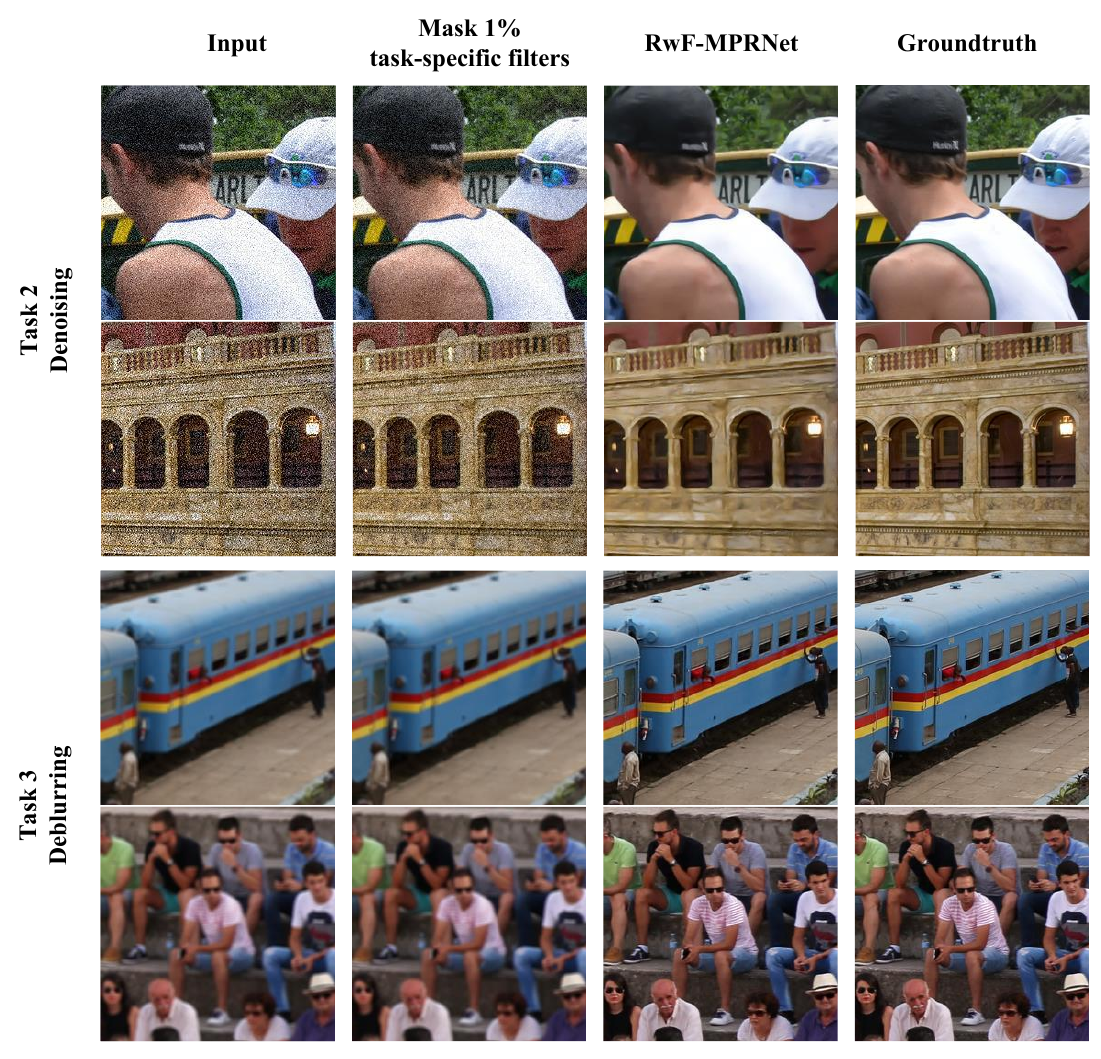}
    \vspace{-4pt}
    \caption{Restored images after resetting 1\% of the backbone's filters, localized for each task, to the shared baseline.}
    \label{fig:mask}
    \vspace{-4pt}
\end{figure*}

A key observation of our work is that the knowledge a backbone acquires for a degradation is concentrated in a small set of filters. To verify it, we take a trained per-task model, reset a subset of its filters to the baseline $\bar\theta$ used for localization, and re-evaluate it in Tab.~\ref{tab:knockout_fraction} and Fig.~\ref{fig:mask}.

\smallskip\noindent\textbf{Resetting the localized filters removes the restoration capability.}
The localized filters of each task cover only $2.05\%$--$2.89\%$ of the backbone's filters. Resetting them reduces PSNR by $12.18$~dB on dehazing, $9.01$~dB on desnowing and $11.54$~dB on LLIE, which removes $99.5\%$--$109.5\%$ of the gain of each task and brings the model to the level of the degraded input. The result is consistent across six independent localizations of LLIE ($11.62\pm1.10$~dB). Moreover, the output becomes almost identical to the degraded input (e.g., $44.51$~dB PSNR against the hazy input, compared with $15.88$~dB for the intact model), i.e., the model stops restoring rather than producing artifacts.

\smallskip\noindent\textbf{Only $1\%$ of the filters is enough.}
As shown in Tab.~\ref{tab:knockout_fraction}, resetting only $1\%$ of the filters already reduces PSNR by $11.4$~dB on dehazing, $7.7$~dB on desnowing and $7.1$~dB on LLIE, while resetting the same number of random filters costs much less, e.g., $0.6$~dB against $4.4$~dB at $0.1\%$ on dehazing.
In Fig.~\ref{fig:mask}, we randomly select 1\% of the generated filters and replace them with the corresponding filters from the baseline. For both new tasks, when 1\% of the generated task-specific filters were masked, the network loses its functionality on the respective new tasks. This observation emphasizes the critical effectiveness of these generated task-specific filters in restoring the corresponding degradation.

\begin{table}[!t]
\centering
\caption{PSNR (dB) after resetting a fraction of the backbone's filters to the shared baseline. ``Localized'' resets filters localized for the task and ``Random'' resets the same number of randomly selected filters (evaluated on $250$/$150$/$100$ test images).}
\resizebox{0.85\linewidth}{!}{
\begin{tabular}{l|cc cc cc}
\toprule
\multicolumn{1}{c|}{\multirow{2}{*}{Filters reset}} & \multicolumn{2}{c}{Dehazing} & \multicolumn{2}{c}{Desnowing} & \multicolumn{2}{c}{LLE} \\
\cmidrule(lr){2-3} \cmidrule(lr){4-5} \cmidrule(lr){6-7}
\multicolumn{1}{c|}{} & Localized & Random & Localized & Random & Localized & Random \\
\midrule
None (intact) & \multicolumn{2}{c}{27.52} & \multicolumn{2}{c}{27.85} & \multicolumn{2}{c}{20.23} \\
\midrule
0.10\% & 23.15 & 26.95 & 26.34 & 27.75 & 19.29 & 20.76 \\
0.25\% & 21.15 & 24.01 & 24.08 & 27.47 & 19.71 & 21.31 \\
0.50\% & 19.40 & 23.72 & 21.20 & 27.23 & 17.13 & 21.83 \\
1.00\% & 16.10 & 18.79 & 20.19 & 23.73 & 13.16 & 18.01 \\
1.50\% & 15.85 & 18.11 & 20.42 & 20.67 & 11.62 & 16.89 \\
All localized & 15.48 & 18.74 & 19.00 & 20.72 & 8.72 & 15.47 \\
\midrule
Degraded input & \multicolumn{2}{c}{15.55} & \multicolumn{2}{c}{18.94} & \multicolumn{2}{c}{9.72} \\
\bottomrule
\end{tabular}}
\label{tab:knockout_fraction}
\end{table}

\smallskip\noindent\textbf{The remaining filters stay unchanged.}
Conversely, all convolution weights outside the localized filters stay bit-exact over the whole task sequence ($94.8\%$ of the backbone), which is why RwF does not forget. Note that these experiments show that a task relies on few filters, not that different tasks use disjoint filters: resetting the filters localized for another task can also degrade a task.

\section{Implementation Details}
\label{app:imple}

\smallskip\noindent\textbf{Benchmark settings.}
We construct a lightweight benchmark from DIV2K~\citep{Agustsson_2017_CVPR_Workshops}, containing 800 training, 100 validation, and 100 test image pairs. We synthesize three degradation types: rain streaks, Gaussian noise with $\sigma \in \{15,25,50\}$, and Gaussian blur with a $25\times25$ kernel and $\sigma=1.6$. To create rain streaks, we generate random noise of the same size as the image and retain the noise within a randomized intensity range, apply an affine transformation to obtain a motion kernel with an angle in $[-45^\circ, 45^\circ]$ and a length of 50, blur the kernel with a Gaussian to obtain streak widths of 1 to 5, and convolve the noise with the kernel. For the five-task extension in Tab.~\ref{table:5tasks}, haze is synthesized with scattering coefficients in $[0.05, 0.2]$ and global atmospheric light in $[0.9, 1]$, and low light with gamma adjustments in $[1.5, 2]$. RwF is combined with MPRNet throughout these experiments.

\smallskip\noindent\textbf{Baselines.}
We compare against sequential fine-tuning (SFT), which uses no anti-forgetting mechanism; regularization-based methods EWC~\citep{kirkpatrick2017overcoming}, LwF~\citep{li2017learning}, and ELI~\citep{joseph2022energy}; replay-based methods iCaRL~\citep{rebuffi2017icarl} and SOIF~\citep{sun2023regularizing}; and PiggybackGAN~\citep{zhai2020piggyback}, a continual conditional image generation method. We additionally evaluate ELI combined with EWC, LwF, and iCaRL.

\smallskip\noindent\textbf{Average forgetting.}
Fig.~\ref{fig:bubble} reports average forgetting~\citep{qu2021recent} over previously learned tasks:
\begin{equation*}
F_q = \frac{1}{q-1}\sum_{p=1}^{q-1} f_p^q,
\end{equation*}
where $q$ is the number of learned tasks and $f_p^q$ is the difference between the highest performance achieved on task $p$ during continual learning and its performance after learning $q$ tasks. Lower $F_q$ indicates less forgetting.


\smallskip\noindent\textbf{Warm-up.}
For each new task ($t \geq 2$), the whole backbone is first trained for 1k iterations. During the first 500 iterations, the model is trained with $\ell_1$ and FFT losses while the prototype of the current task is initialized by EMA; during the remaining 500 iterations, the prototypical contrastive loss is added, using the new prototype and all stored ones.


\begin{algorithm}[!t]
\caption{Pseudocode of RwF in a PyTorch-like style.}
\label{alg:rwf}
\definecolor{codeblue}{rgb}{0.25,0.5,0.5}
\definecolor{codekw}{rgb}{0.85,0.18,0.50}
\lstset{
  backgroundcolor=\color{white},
  basicstyle=\fontsize{7.5pt}{7.5pt}\ttfamily\selectfont,
  columns=fullflexible,
  breaklines=true,
  captionpos=b,
  commentstyle=\fontsize{7.5pt}{7.5pt}\color{codeblue},
  keywordstyle=\fontsize{7.5pt}{7.5pt}\color{codekw},
}
\begin{lstlisting}[language=python]
# f(theta, x) -> (y_hat, e): backbone, outputs the restored image and an embedding
# theta0: pre-trained model of task 1; D[t] = (X[t], Y[t]): data of task t
# L: localized filters; B: filter bank, one filter per row; P[t]: prototype of task t
# W1, W2, w: trainable weights, re-initialized for each task
# L, B, W1, W2 and w are kept per layer; the layer index is omitted
 
theta_bar = finetune(theta0, X[1], X[1])  # IG baseline (self-reconstruction)
theta, B = theta0, []
P[1] = ema(emb(theta0, X[1]))
 
for t in range(2, T + 1):
    theta_warm, P[t] = warmup(theta, D[t])  # full fine-tuning (Sec. 3.1)
 
    # filter localization with integrated gradients (Sec. 3.2)
    ig = (theta_warm - theta_bar) * mean_grad(theta_bar, theta_warm, D[t])
    s = ig.abs().flatten(1).sum(1)  # IG score of each filter
    L = select(s)  # layer- then filter-level, Eq. (4)-(7)
    L = expand(L, theta_warm)  # filter diversity expanding, Eq. (8)
 
    # continual filter learning (Sec. 3.3)
    B_prev = B
    B = cat([B_prev, theta_warm[L]])
    for x, y in loader(D[t]):
        g = mm(mm(B.T, W1), W2).T  # generate filters from the bank
        if len(B_prev) >= len(g):
            g = g + 0.1 * attn(g, B_prev)  # cross-task attention, Eq. (13)-(14)
        lam = sigmoid(w)
        theta[L] = (1 - lam) * theta_warm[L] + lam * g
        y_hat, e = f(theta, x)
        loss = l1(y_hat, y) + 0.1 * l_proto(e, P) + 0.01 * l_fft(y_hat, y)  # Eq. (18)
        loss.backward()
        update(W1, W2, w)
        P[t] = 0.9 * P[t] + 0.1 * e.mean(0)
 
    B[-len(g):] = theta[L]  # keep the learned filters for later tasks
    save(t, L, theta[L])
 
\end{lstlisting}
\end{algorithm}

\smallskip\noindent\textbf{Continual filter learning.}
After localization, all weights are restored to $\theta^{t-1}$ except the localized filters, which keep their warmed-up values. For every selected convolution, these filters are appended to its filter bank $\mathcal{B}$, and a rank-8 generator $(W_1, W_2)$ with a learnable mixing weight $\lambda$ is created. The localized filters are then replaced by $(1-\lambda)\,\theta + \lambda\,\tilde{\theta}$, where $\tilde{\theta}$ is generated from the filter bank (Alg.~\ref{alg:rwf}). Only $W_1$, $W_2$ and $\lambda$ are trained, for 150k iterations with the loss $10\,\ell_1 + \mathcal{L}_{\mathrm{FFT}} + 0.1\,\mathcal{L}_{\mathrm{proto}}$, AdamW, a constant learning rate and gradient clipping at 1.0, while the rest of the backbone stays frozen. After training, the generated filters are written into the backbone, so each task is deployed as a plain HINT model and only its localized filters need to be stored.

\smallskip\noindent\textbf{Cross-task attention.}
When generating filters for a convolution whose bank holds at least as many filters as are being generated, the generated filters attend to the bank: queries and keys are projected to $\min(32, d/2)$ dimensions, values keep the filter dimension $d$, and the attention output is added to the generated filters with a weight of $0.1$. In our implementation, these projections keep their random initialization and are not updated during training. Since the generated filters are written into the backbone after each task, the attention brings no extra computation at inference.

\smallskip\noindent\textbf{Prototypical contrastive learning} operates on features extracted from bottleneck layers. Specifically, we tap into the deepest encoder stage of the backbone, yielding a feature tensor of shape (B, C$_b$, H', W') where C$_b$=256 is the bottleneck channel dimension. A global average pooling layer then reduces this tensor to a compact (B, C$_b$) embedding, and both the EMA prototype update and the contrastive objective are computed. This design ensures that prototypes capture degradation-specific representations at a semantically rich abstraction level, rather than operating in the low-dimensional RGB output space where different restoration tasks would be largely indistinguishable, since all tasks converge toward similar clean-image statistics in the output domain.

\smallskip\noindent\textbf{Algorithm.}
Alg.~\ref{alg:rwf} summarizes RwF in a PyTorch-like style. The first task uses the released model and defines the IG baseline. For each new task, the backbone is first warmed up on the new data, and the key filters are localized in a coarse-to-fine manner and expanded with a few dissimilar filters. Then, all other weights are restored, and only a small generator per selected convolution is trained to produce the key filters from the filter bank. Finally, the generated filters are written into the backbone and stored together with their positions, and the model waits for the next task.

\smallskip\noindent\textbf{Task identification at inference.}
Since each task is served by its own localized filters, the model needs the task identity of a test image. Instead of requiring it as an input, we infer it with a lightweight prototype-based router. The degraded image, resized to $256\times256$, is passed through the encoder of the frozen first-task model, and its global-average-pooled latent feature ($384$-d) is used as the descriptor. For each task, a Gaussian with a shrinkage covariance is fitted to the descriptors of its training images and serves as the task prototype; a test image is assigned to the task with the highest likelihood, and the filters of that task are assembled into the backbone. The router requires no training and stores only a mean and a covariance per task. As shown in Tab.~\ref{tab:routing}, routing is accurate on all restoration tasks. 

\begin{table}[!t]
\centering
\caption{Task identification at inference on the four-task public benchmark. ``PSNR change'' is the PSNR obtained with the task predicted by the prototype router minus the PSNR obtained with the ground-truth task identity.}
\resizebox{0.7\linewidth}{!}{
\begin{tabular}{l|ccccc}
\toprule
Metric & Deraining & Dehazing & Desnowing & LLE & Average \\
\midrule
Routing accuracy (\%) & 98.0 & 90.4 & 100.0 & 100.0 & 97.1 \\
\midrule
PSNR change (dB) & $-0.10$ & $-1.18$ & $0.00$ & $0.00$ & $-0.32$ \\
\bottomrule
\end{tabular}}
\label{tab:routing}
\end{table}

\begin{table}[t]
\centering
\caption{Filter consumption and parameter overhead when the number of tasks grows to ten. All numbers are percentages.}
\resizebox{0.7\linewidth}{!}{
\begin{tabular}{l|ccccc}
\toprule
\multicolumn{1}{c|}{\multirow{2}{*}{Statistic}} & \multicolumn{5}{c}{Number of learned tasks} \\
\cmidrule(lr){2-6}
\multicolumn{1}{c|}{} & 2 & 4 & 6 & 8 & 10 \\
\midrule
Filters localized by the new task & 2.05 & 2.44 & 2.87 & 2.95 & 2.67 \\
Reused from previous tasks & 0.0 & 54.1 & 71.9 & 81.1 & 88.3 \\
\midrule
Edited filters, random selection~$\downarrow$ & 2.05 & 6.71 & 10.49 & 13.43 & 15.38 \\
Edited filters, RwF (ours)~$\downarrow$ & 2.05 & 5.25 & 6.83 & 7.77 & \textbf{8.35} \\
\midrule
Stored params, separate models~$\downarrow$ & 100 & 300 & 500 & 700 & 900 \\
Stored params, RwF (ours)~$\downarrow$ & 2.66 & 10.02 & 17.29 & 25.40 & \textbf{33.07} \\
\bottomrule
\end{tabular}}
\label{tab:scalability}
\end{table}

\section{Additional Experimental Analysis}
\label{sec:exp}

\label{app:scalability}
\smallskip\noindent\textbf{Capacity analysis.}
To test whether RwF remains effective as tasks accumulate, we extend the four-task benchmark of Sec.~\ref{sec:pub} to ten tasks on HINT by appending deblurring and denoising with $\sigma\in\{10,15,25,35,50\}$ (tasks 6--10 are trained for only 10k iterations); Tab.~\ref{tab:scalability} summarizes the statistics. Each new task localizes $2.05\%$--$2.95\%$ of the backbone's filters regardless of its position in the sequence, and later tasks mostly reuse filters localized before ($88.3\%$ reuse at the tenth task), so the filters edited by all ten tasks cover only $8.35\%$ of the backbone, against $15.38\%$ for a random selection of the same size per layer; no layer is exhausted, the most occupied one (among layers with at least 32 filters) using $60.4\%$ of its filters. Since each task stores its own values of the localized filters plus its adapter, the stored parameters grow linearly: after ten tasks RwF stores $33.1\%$ of the backbone in total, i.e., $3.7\%$ per task, comparable to a rank-8 LoRA ($3.5\%$) and far below one model per task ($900\%$). Meanwhile, all $22.0$M convolution and linear weights outside the edited filters ($88.6\%$ of the backbone) stay unchanged from the first task to the tenth, so the performance of every learned task is preserved no matter how many tasks follow.

\begin{figure*}[!t]
    \centering
    \includegraphics[width=0.85\textwidth]{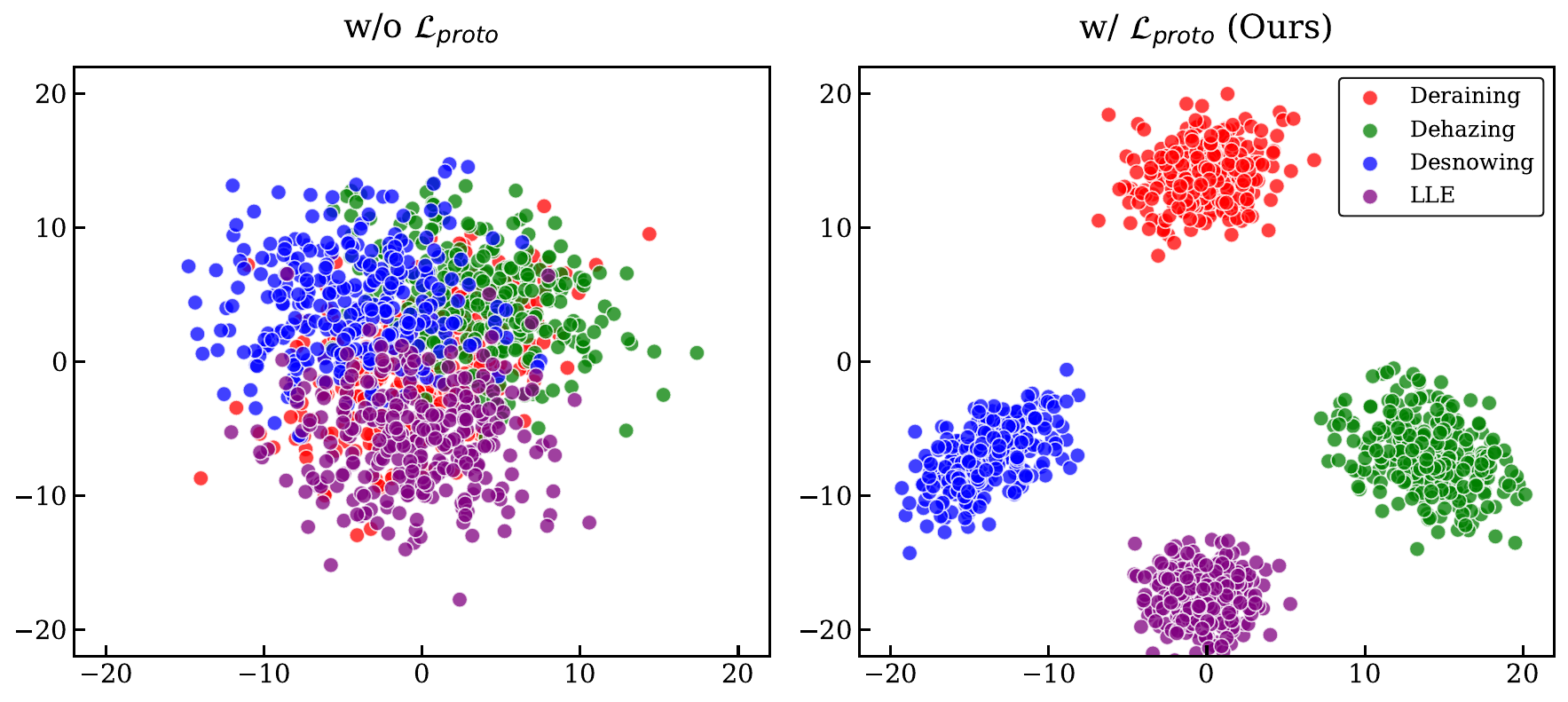}
    \vspace{-4pt}
    \caption{t-SNE visualization of latent feature distributions across four restoration tasks, comparing models trained with and without prototypical contrastive learning.}
    \label{fig:tsne}
    \vspace{-4pt}
\end{figure*}
\smallskip\noindent\textbf{Effect of Prototypical Contrastive Learning}
To intuitively demonstrate the efficacy of the proposed prototypical contrastive loss $\mathcal{L}_{proto}$, we visualize the latent feature representations of different degradation tasks using t-SNE~\citep{van2008visualizing}, as shown in Fig.~\ref{fig:tsne}. Without $\mathcal{L}_{proto}$, the feature distributions of distinct tasks (e.g., deraining, dehazing, and desnowing) severely overlap, which inevitably leads to gradient interference and catastrophic forgetting during sequential training. In contrast, by pulling features toward their respective task prototypes and pushing them away from historical ones, our RwF framework effectively enforces task-discriminative representations by prototypical contrastive learning. The resulting latent space exhibits well-separated clusters for each degradation pattern, explicitly mitigating negative transfer across sequential tasks.

\begin{figure*}[!t]
    \centering
    \includegraphics[width=1\textwidth]{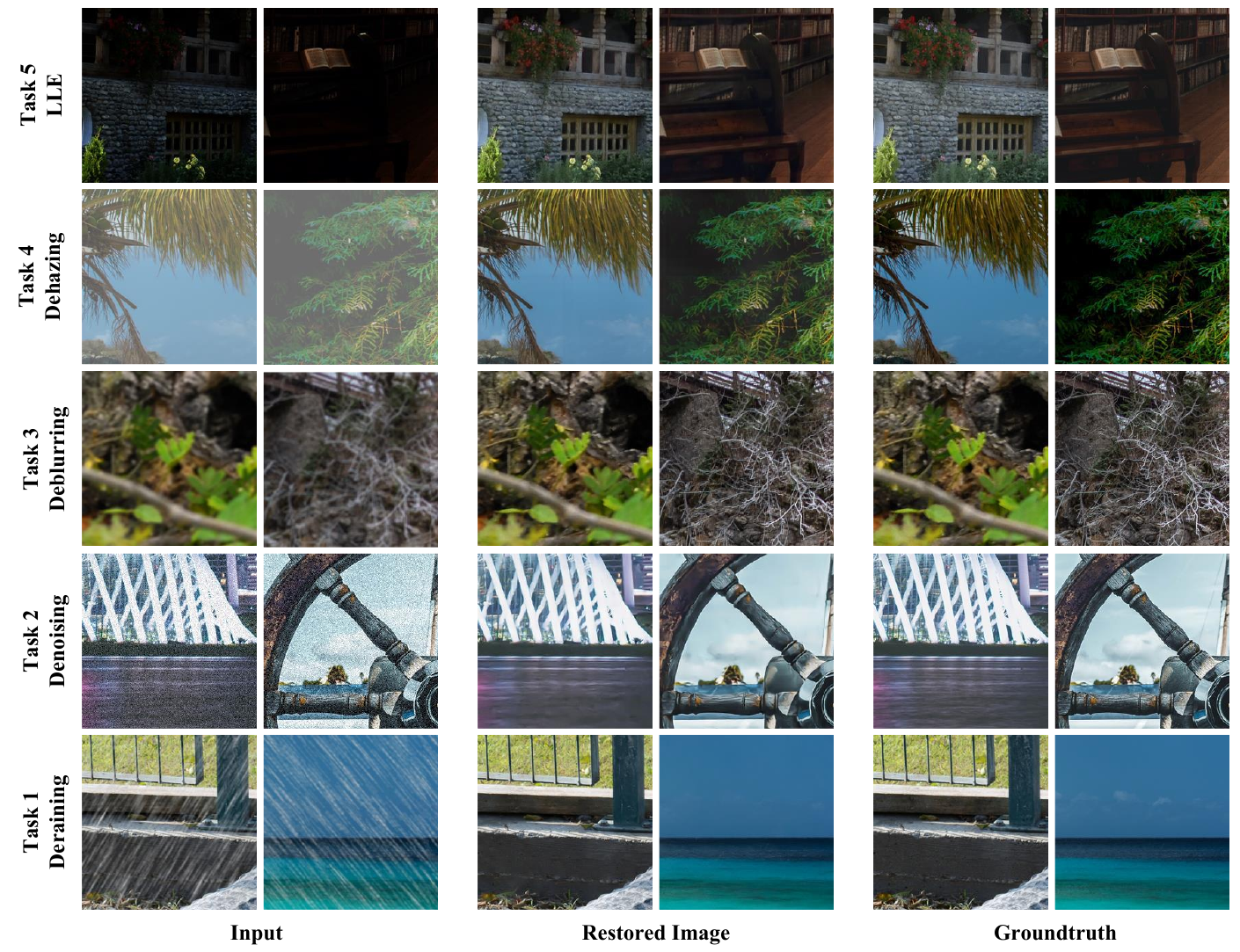}
    \vspace{-4pt}
    \caption{Visualization of restored images on randomly selected samples from test set after expanding to 5 restoration tasks in preliminary experiments, with the training sequence of deraining, denoising, deblurring, dehazing and LLE.}
    \label{fig:5tasks}
    \vspace{-4pt}
\end{figure*}

\smallskip\noindent\textbf{Preliminary experiments with more tasks.}
To further validate the scalability and robustness of our proposed RwF framework over longer continual learning sequences, we extend the preliminary experiments from original three tasks to a five-task sequence. Specifically, we sequentially introduce dehazing and low-light enhancement following the initial deraining, denoising, and deblurring tasks. As demonstrated in Tab~\ref{table:5tasks} and Fig~\ref{fig:5tasks}, RwF maintains its high-fidelity restoration performance across all five tasks. This demonstrates its superior plasticity and stability even under extended and more complex continual restoration scenarios.

\begin{table}[!t]
\centering
\caption{Quantitative results after expanding to 5 restoration tasks, with the training sequence of deraining, denoising, deblurring, dehazing and LLE.}
\resizebox{\linewidth}{!}{
\begin{tabular}{l|cc cc cc cc cc cc}
\toprule
\multirow{2}{*}{Method} & \multicolumn{2}{c}{Deraining} & \multicolumn{2}{c}{Denoising} & \multicolumn{2}{c}{Deblurring} & \multicolumn{2}{c}{Dehazing} & \multicolumn{2}{c}{LLE} & \multicolumn{2}{c}{Average} \\
\cmidrule(lr){2-3} \cmidrule(lr){4-5} \cmidrule(lr){6-7} \cmidrule(lr){8-9} \cmidrule(lr){10-11} \cmidrule(lr){12-13}
 & PSNR & SSIM & PSNR & SSIM & PSNR & SSIM & PSNR & SSIM & PSNR & SSIM & PSNR & SSIM \\
\midrule
SFT & 13.64 & 0.633 & 14.85 & 0.521 & 15.60 & 0.627 & 11.70 & 0.713 & \textbf{28.75} & \textbf{0.939} & 16.91 & 0.687 \\
PiggybackGAN~\citep{zhai2020piggyback} & 32.51 & 0.895 & 29.54 & 0.824 & 31.46 & 0.878 & 27.80 & 0.867 & 27.45 & 0.882 & 29.75 & 0.869 \\
RwF-MPRNet (Ours) & \textbf{38.52} & \textbf{0.942} & \textbf{33.69} & \textbf{0.874} & \textbf{34.49} & \textbf{0.913} & \textbf{32.01} & \textbf{0.920} & 28.08 & 0.916 & \textbf{33.36} & \textbf{0.913} \\
\bottomrule
\end{tabular}}
\vspace{-12pt}
\label{table:5tasks}
\end{table}

\smallskip\noindent\textbf{Perceptual quality.}
Tab.~\ref{tab:lpips} reports LPIPS~\citep{zhang2018unreasonable} and DISTS~\citep{ding2022image} to compare perceptual quality between competing methods, where both are consistent with PSNR and SSIM. SFT severely degrades the perceptual quality of the earlier tasks, e.g., the LPIPS of deraining rises to $0.327$ after the last task, which is even worse than the degraded input ($0.271$), whereas RwF-HINT keeps it at $0.011$. RwF-HINT achieves the lowest average LPIPS among the HINT-based methods without storing any past data, while SFT with rehearsal needs to reuse training images per task to reach a similar level. On LLE, the last task, SFT reaches a lower LPIPS since it is fine-tuned without any constraint. For DiffUIR, RwF reduces the average LPIPS of the zero-shot model from $0.104$ to $0.083$.

\begin{table}[!t]
\centering
\caption{LPIPS and DISTS (lower is better) on the four-task public benchmark with the order of deraining, dehazing, desnowing, and low-light enhancement (LLE), evaluated on all tasks after learning the last task. Best results within each backbone are in bold.}
\resizebox{\linewidth}{!}{
\begin{tabular}{l|cccccccccc}
\toprule
\multicolumn{1}{c|}{\multirow{2}{*}{Method}} & \multicolumn{2}{c}{Deraining} & \multicolumn{2}{c}{Dehazing} & \multicolumn{2}{c}{Desnowing} & \multicolumn{2}{c}{LLE} & \multicolumn{2}{c}{Average} \\
\cmidrule(lr){2-3}\cmidrule(lr){4-5}\cmidrule(lr){6-7}\cmidrule(lr){8-9}\cmidrule(lr){10-11}
\multicolumn{1}{c|}{} & LPIPS & DISTS & LPIPS & DISTS & LPIPS & DISTS & LPIPS & DISTS & LPIPS & DISTS \\
\midrule
Degraded input & 0.271 & 0.182 & 0.104 & 0.123 & 0.356 & 0.209 & 0.519 & 0.352 & 0.313 & 0.216 \\
\midrule
DiffUIR (zero-shot)~\citep{zheng2024selective} & 0.120 & 0.097 & \textbf{0.008} & \textbf{0.014} & 0.147 & 0.101 & \textbf{0.139} & \textbf{0.125} & 0.104 & 0.084 \\
SFT-DiffUIR~\citep{zheng2024selective} & 0.272 & 0.183 & 0.076 & 0.078 & 0.270 & 0.170 & 0.165 & 0.139 & 0.196 & 0.143 \\
\textbf{RwF-DiffUIR} & \textbf{0.033} & \textbf{0.038} & \textbf{0.008} & \textbf{0.014} & \textbf{0.145} & \textbf{0.100} & 0.147 & 0.126 & \textbf{0.083} & \textbf{0.069} \\
\midrule
SFT-HINT~\citep{zhou2025devil} & 0.327 & 0.217 & 0.272 & 0.209 & 0.405 & 0.232 & \textbf{0.142} & \textbf{0.124} & 0.287 & 0.195 \\
SFT + Rehearsal-HINT & 0.034 & 0.046 & 0.018 & 0.025 & 0.148 & \textbf{0.111} & 0.156 & 0.136 & 0.089 & \textbf{0.079} \\
\textbf{RwF-HINT} & \textbf{0.011} & \textbf{0.018} & \textbf{0.017} & \textbf{0.024} & \textbf{0.143} & 0.113 & 0.172 & 0.159 & \textbf{0.086} & \textbf{0.079} \\
\bottomrule
\end{tabular}}
\label{tab:lpips}
\end{table}

\begin{figure*}[!t]
    \centering
    \includegraphics[width=1\textwidth]{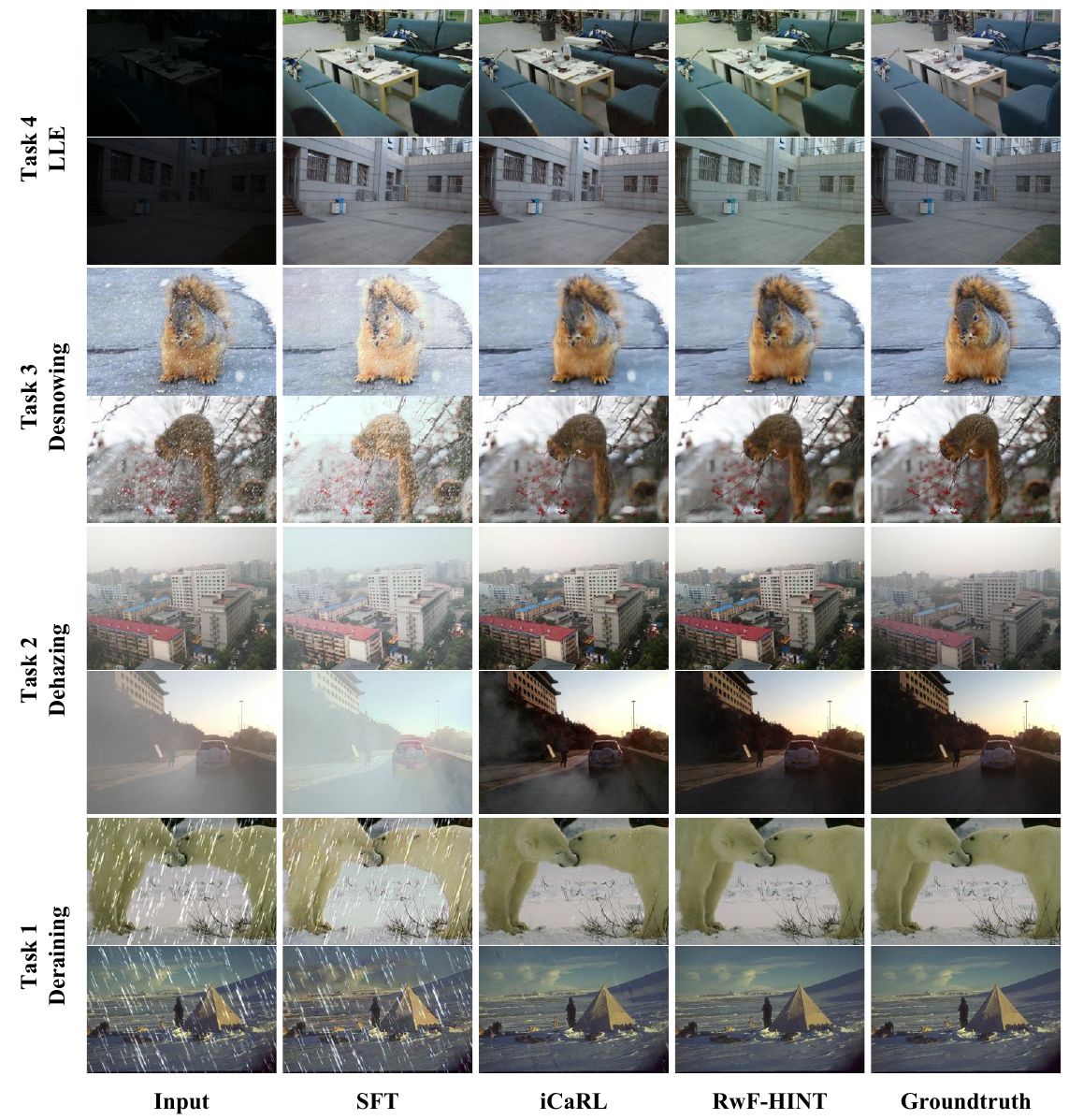}
    \vspace{-4pt}
    \caption{Visualization of restored images on randomly selected samples from test set of four-task public benchmarks.}
    \label{fig:m1}
    \vspace{-4pt}
\end{figure*}
\begin{figure*}[!t]
    \centering
    \includegraphics[width=1\textwidth]{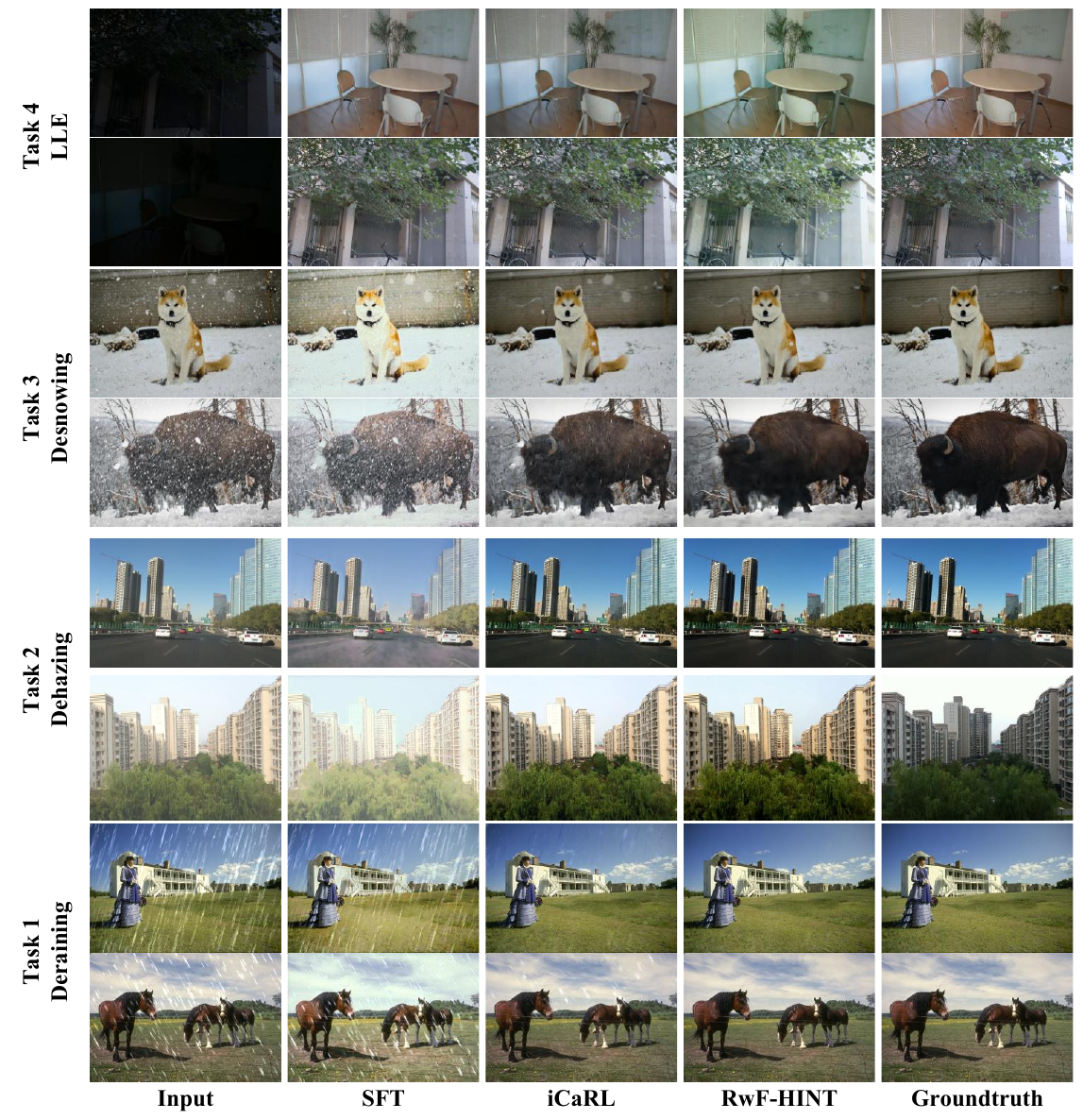}
    \vspace{-4pt}
    \caption{Visualization of restored images on randomly selected samples from test set of four-task public benchmarks.}
    \label{fig:m2}
    \vspace{-4pt}
\end{figure*}

\smallskip\noindent\textbf{More Visualizations of main experiments.}
Due to space constraints in the main manuscript, we provide more comprehensive qualitative comparisons on the four-task public restoration benchmark (Rain100L, SOTS, Snow100K, and LOLv2-Real) in Fig.~\ref{fig:m1} to Fig.~\ref{fig:m2}. As consistently observed, sequential fine-tuning (SFT) and iCaRL often produce severe blending artifacts and color shifts when evaluated on previously learned tasks. Conversely, our RwF seamlessly recovers sharp structural edges, rich textures, and natural color distributions across all diverse degradations.

\end{document}